\pdfoutput=1

\documentclass[11pt]{article}

\usepackage{acl}

\usepackage{times}
\usepackage{latexsym}

\usepackage[T1]{fontenc}

\usepackage[utf8]{inputenc}

\usepackage{microtype}

\usepackage{graphicx}
\usepackage{subfigure}
\usepackage{booktabs} 
\usepackage{multicol}
\usepackage{multirow}
\usepackage{amsmath,amsthm,amssymb,amsxtra} 
\usepackage{siunitx} 
\usepackage{tabularx}

\usepackage{xcolor}

\definecolor{redpink}{rgb}{ .898,  .263,  .875}

\title{Hint-before-Solving Prompting: Guiding LLMs to \\Effectively Utilize Encoded Knowledge}

\author{Jinlan Fu\textsuperscript{1}\thanks{\ \  These two authors contributed equally.}, 
Shenzhen Huangfu\textsuperscript{2}$^*$, 
Hang Yan\textsuperscript{3}, 
See-Kiong Ng\textsuperscript{1},
Xipeng Qiu\textsuperscript{2,3}\thanks{\  \ Corresponding author. }, 
\\
  \textsuperscript{1}National University of Singapore, 
  \textsuperscript{2}School of Computer Science, Fudan University \\
  \textsuperscript{3}Shanghai AI Laboratory\\
  \texttt{\small{\{jinlanjonna, shenzhenhuangfu\}@gmail.com,}} \\
  \texttt{\small{yanhang@pjlab.org.cn, seekiong@nus.edu.sg, xpqiu@fudan.edu.cn}}\\ 
}

\begin{document}
\maketitle
\begin{abstract}

Large Language Models (LLMs) have recently showcased remarkable generalizability in various domains. 
Despite their extensive knowledge, LLMs still face challenges in efficiently utilizing encoded knowledge to develop accurate and logical reasoning processes. 
To mitigate this problem, we introduced Hint-before-Solving Prompting (HSP), which guides the model to generate hints (e.g., specific knowledge or key ideas)  for solving the problem and then generate solutions containing intermediate reasoning steps. 
Since HSP is orthogonal to prompting methods (e.g., Chain-of-Thought (CoT)), we applied HSP to CoT, Least-to-Most, Plan-and-Solve, and Standard promptings.
The results of extensive experiments on 6 reasoning benchmarks and 4 open-source LLMs demonstrate that HSP can effectively improve the accuracy of reasoning tasks:
(1) By applying high-quality hint-enhanced HSP to CoT prompting, Llama2-70B-Chat shows an improvement of 9.7.
(2) Beyond exploring training-free LLM capabilities, we built the HSPMATH dataset based on HSP and fine-tuned Llemma-7B, reaching 64.3 accuracy, surpassing GPT-3.5 and WizardMath-13B.
We make our code and dataset publicly available at \url{https://github.com/jinlanfu/HSP}.
\end{abstract}

\section{Introduction}

Benefiting from extensive training corpora and computational resources, Large Language Models (LLMs) have reached state-of-the-art performance in numerous Natural Language Processing (NLP) tasks~\cite{touvron:llama1,openai:gpt4,touvron:llama2,xin:llm_survey,mistral2023mixtral}. 
However, LLMs still face challenges in complex reasoning tasks, such as mathematical reasoning~\cite{lu:mathreason,wizardmath:luo,mathprompter:shima}  and commonsense reasoning~\cite{common:bhargavi,commonnlp:maarten}.
Although possessing a wealth of knowledge, LLMs always fail to accurately apply encoded knowledge to generate coherent and strongly logical reasoning chains when addressing reasoning tasks.

\begin{figure}[!t]
    \centering
    \includegraphics[width=0.98\linewidth]{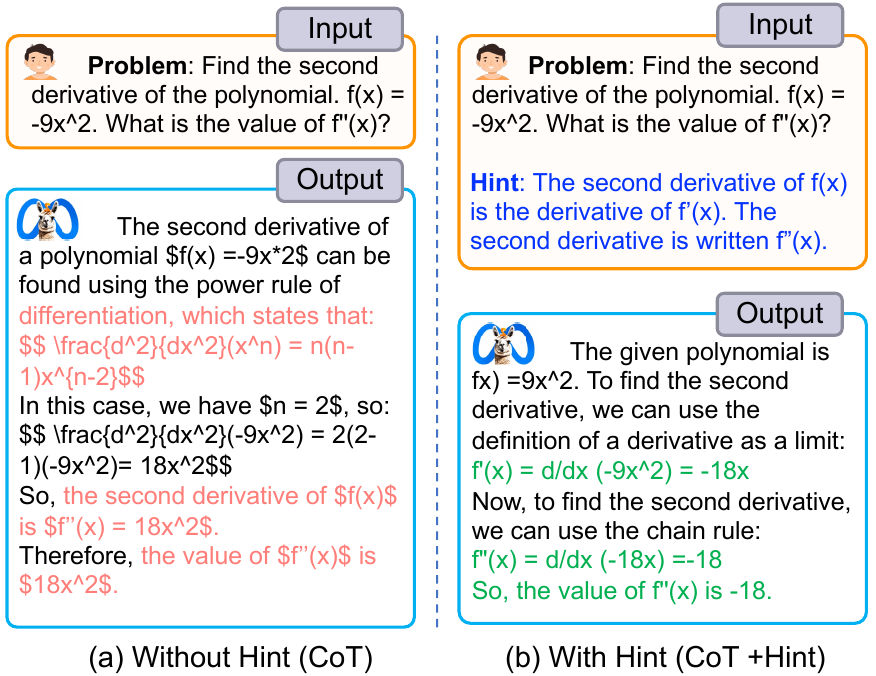}
    \vspace{-8pt}
    \caption{
    The output comparison of Llama-2-Chat-70B solving a math problem (calculus) with and without a hint. 
    Red text indicates erroneous information; green text indicates correct reasoning.
    Findings:
    (1) having a hint can help the LLM understand the problem.
    (2) The LLM possesses knowledge of calculus, and with a hint, it can accurately apply this knowledge.
}
	\label{fig:hint_example}
\end{figure}

To improve the performance of LLMs on complex reasoning tasks, existing works have made several attempts. 
These previous works include fine-tuning on complete training datasets~\cite{wizardmath:luo,metamath:yu,mammoth:yue} , training-free methods based on prompt engineering~\cite{zhou:ltm,wang:psp,fu:complexitycot,faithfulcot:lyu,ZhaoLJQB23}, or enhancing by retrieving knowledge from external knowledge bases~\cite{yao:react,reretrival:he,leanDojo:yang}.
Supervised fine-tuning methods are resource-intensive; current prompt engineering seldom attempt to improve LLMs' ability to use accurate knowledge; retrieval augmentation methods are limited to specific tasks. For example, mathematical reasoning that includes many special symbols is difficult to access relevant knowledge through keyword or semantic retrieval.

\begin{figure}[!t]
    \centering
    \includegraphics[width=0.8\linewidth]{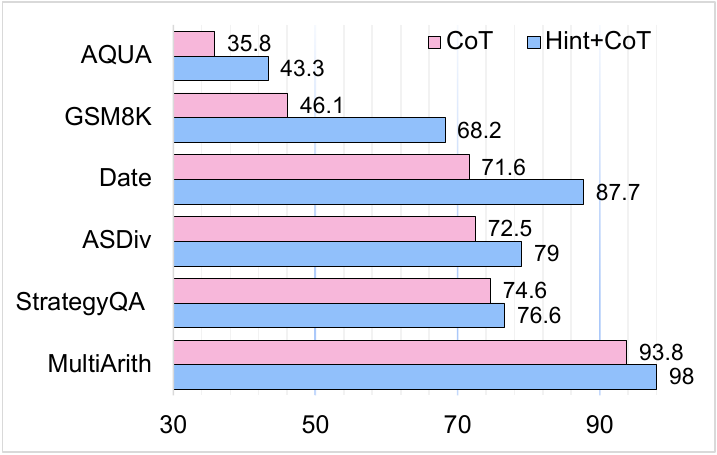}
    \vspace{-8pt}
    \caption{Results for Llama-2-Chat-70B (under CoT prompting) with or without introducing high-quality hints across six reasoning datasets. Findings: introducing hints lead to significant improvements, with an average relative increase of 9.7\%.}
	\label{fig:examp_res}
\end{figure}

To mitigate these problems, in this work, we explore how LLMs can effectively utilize their encoded knowledge to enhance their reasoning logic and accuracy.
We found that providing LLMs with hints effectively guides their use of encoded knowledge for problem-solving. Fig.~\ref{fig:hint_example} illustrates this by comparing Llama2-70B's outputs on a \textit{calculus problem} with and without hints.
The LLM cannot utilize \textit{calculus knowledge} to solve the problem without any hints, as shown in Fig.\ref{fig:hint_example}-(a).
However, when given a hint (as shown in Fig.\ref{fig:hint_example}-(b)): ``\textit{... The second derivative is written $f''(x)$.}'' the LLM can accurately apply its ``\textit{calculus knowledge}'' to generate a correct and logical solution with intermediate reasoning.
The reason can be attributed to that the hint suggested that  ``$f''(x)$ denotes the second derivative'', which helped the LLM to better understand the target of the problem.
Moreover, we conducted quantitative analysis on six reasoning datasets by introducing hints generated by GPT-4. 
The experimental results are shown in Fig.~\ref{fig:examp_res}. We can find that giving high-quality hints can effectively improve reasoning performance.

However, it is challenging to provide high-quality hints for every sample. 
To address this problem, we propose the Hint-before-Solving (HSP) prompting method, which allows LLMs to generate hints on their own before solving a problem.
The hints may include knowledge necessary for solving the problem (e.g., the hint shown in Fig.~\ref{fig:hint_example}-(b)), analyzing the question, and providing essential ideas for the solution.
Our explorations of Hint-before-Solving (HSP) Prompting in this paper are driven by following research questions:

Q1: \textit{Can HSP guiding LLMs to autonomously generate helpful hints be effective?}
To answer this question, we incorporated HSP into four well-performing prompting methods to investigate how HSP performs (EXP-I). Furthermore, we examined the effectiveness of the HSP variant, HSP2, which provides hints and solutions in two stages (EXP-II). And explore the upper bound of LLMs under the HSP2 framework (EXP-III). (Sec.~\ref{sec:can_work})

Q2: \textit{Does HSP still work when dealing with tasks that are challenging for LLMs?}
In other words, if a task is difficult for LLMs, can they still provide helpful hints? To answer this question, we evaluated the challenging MATH dataset (EXP-IV). Furthermore, we explore how LLMs perform under the self-consistency setting (EXP-V). (Sec.~\ref{sec:hard_task})

Q3: \textit{How do LLMs perform if they are supervised fine-tuned on a large-scale HSP prompting dataset?}
To answer this question, we constructed the HSPMATH dataset based on GSM8K and conducted supervised fine-tuning on Llemma-7B and Llama-2-13B. The experimental results show that we achieved a performance of 61.7 on Llemma-7B, surpassing GPT3.5. (EXP-VI, Sec.~\ref{sec:sft})

The main contributions of this work are summarized as below:

\noindent
(1) We discovered that providing hints allows LLMs to use their encoded knowledge accurately and effectively. For quantitative analysis, with GPT-4 generated hints, Llama-2-Chat-70B's accuracy increased by nearly 10\% across six datasets.

\noindent
(2) We propose the HSP prompting method, allowing LLMs to automatically generate useful hints. We conducted extensive experiments and analyses on applying HSP to four popular prompting methods to verify HSP's effectiveness.

\noindent
(3) We collected 75,000 samples enhanced with hints, namely HSPMATH (to be released), and fine-tuned Llemma-7B to achieve 64.3 accuracy, surpassing GPT-3.5 (57.1) and WizardMath-13B (63.9).

\begin{figure*}[!th]
    \centering
    \includegraphics[width=0.99\linewidth]{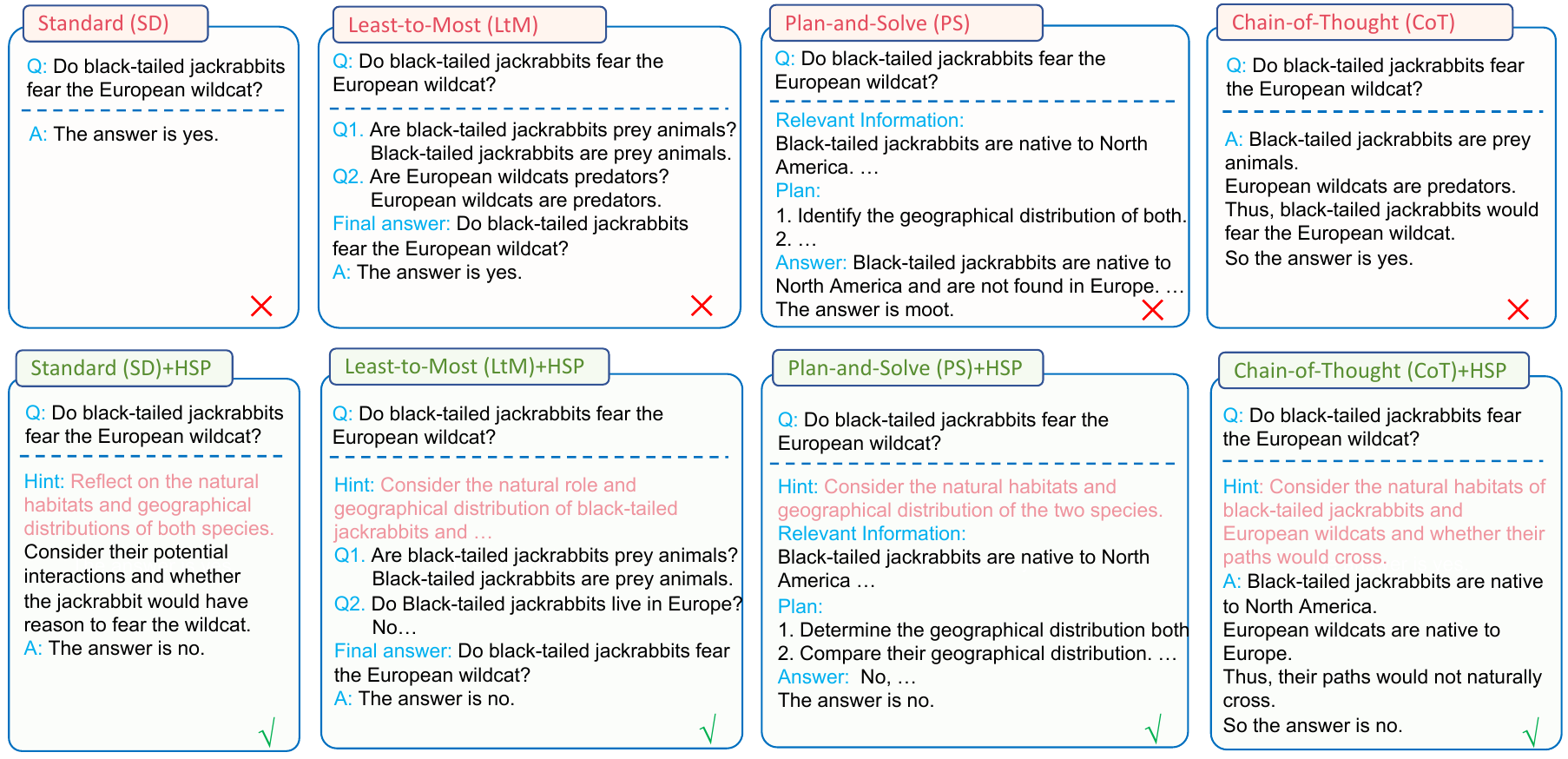}
    \vspace{-8pt}
    \caption{Examples of input and output before (four examples at the top) and after (four examples at the bottom) applying HSP to standard Least-to-Most, Plan-and-Solve, and CoT promptings. The red text in the textbox indicates hints. We find that hints from LLMs, including problem-solving ideas close to the correct answer (e.g., geographical distributions of both species), guide LLMs to use accurate knowledge for correct and logical reasoning.}
	\label{fig:framework}
\end{figure*}

\section{Hint-before-Solving Prompting}
The prominent Chain-of-Thought (CoT)~\cite{wei2022chain} prompting method has inspired various prompting techniques to improve the LLMs' performance.  
Such as Least-to-Most~\cite{zhou2022least}, tree-of-thought~\cite{tree:yao}, graph-oc-thought~\cite{graph:besta}, plan-and-solve prompting~\cite{wang:psp}.
In this work, we aim to design a new prompting method that allows LLMs to better utilize their encoded knowledge, namely Hint-before-Solving Prompting (HSP). HSP enables LLMs to explicitly generate hints for solving problems. The hints can be knowledge or key ideas for solving the problem or analyzing the question, etc., and developing an accurate and logical intermediate reasoning process before predicting the final answer.

HSP can be used in conjunction with some of the existing natural language forms of prompting methods (e.g., CoT). 
Fig.~\ref{fig:framework} shows examples of HSP integrated with four existing prompting methods, namely standard prompting, Least-to-Most prompting~\cite{zhou2022least}, Plan-and-Solve prompting~\cite{wang2023plan}, and Chain-of-Thought prompting~\cite{wei2022chain}. 
We can observe that the current hints provide LLMs with perspectives for thought (e.g., \textit{consider the geographical distribution of black-tailed
jackrabbits ...}), enhancing the effectiveness of prompting methods with the introduction of HSP.

\section{Experiment}

\begin{table}[htb]
  \centering \footnotesize
  \renewcommand\tabcolsep{1.5pt}
    \begin{tabular}{lccccccc}
    \toprule
    Number & G8K & ASDiv & MArith & AQUA &MATH & SQA   & Date \\
    \midrule
    Samples & 1,319  & 2,097  & 596   & 254  &5,000  & 2,290  & 359 \\
    Examples & 8     & 8     & 8     & 8   &4  & 6     & 10 \\
    \bottomrule
    \end{tabular}%
      \caption{The number of test samples and  prompting examples across seven datasets. }
  \label{tab:statistics}%
\end{table}%

\subsection{Large Language Model}
To verify the performance of our proposed method, we consider Mixtral-8x7B-Instruct-v0.1 (\textit{Mix-56B})~\cite{mistral2023mixtral} and Llama-2-Chat~\cite{touvron2023llama} family models, where
Llama-2-Chat-7B (\textit{Lm2-7B}), Llama-2-Chat-13B (\textit{Lm2-13B}), Llama-2-Chat-70B (\textit{Lm2-70B}) were studied. Note, the italicized text in parentheses represents the abbreviated names of the models.

\subsection{Datasets}
We evaluated the effectiveness of HSP across multiple datasets for mathematical and common sense reasoning tasks. Tab.~\ref{tab:statistics} shows the number of test samples for these datasets and the number of samples for prompting in a few-shot setting.

\noindent
\textbf{Mathematical Reasoning}
We considered five popular mathematical reasoning datasets, namely
\textit{GSM8K (G8K)}~\cite{cobbe2021training},
\textit{MultiArith (MArith)}~\cite{MultiArith},
\textit{AQuA}~\cite{aqua},
\textit{ASDiv}~\cite{asdiv}, and
\textit{MATH}~\cite{hendrycks2021measuring}.

\paragraph{Commonsense Reasoning}
Two common sense reasoning datasets were also taken into account, which are
\textit{StrategyQA (SQA)}~\cite{StrategyQA} and
\textit{Date Understanding (Date)}~\cite{date}.

\subsection{Baselines}
\label{sec:baselines}
The baseline Prompting methods considered in this work are listed below:

\noindent
(1)
\textit{Standard Prompting (SD)}~\cite{brown2020language} generates the answer for the given question without intermediate steps.
(2)
\textit{Chain-of-Thought Prompting (CoT)}~\cite{wei2022chain} generate step-by-step solutions to a given problem.
(3)
\textit{Least-to-Most Prompting (LtM)}~\cite{zhou2022least} involves decomposing a complex problem into simple subproblems.
(4)
\textit{Plan-and-Solve Prompting (PS)}~\cite{wang2023plan} aims to handle the multi-step reasoning task by planning and solving each plan target.

To validate the effectiveness of the our HSP, we reimplemented some previous prompting methods.
\textit{To ensure a fair comparison, we did not deliberately reproduce results reported in previous papers but rather aimed to maintain consistency in the experimental setup.
For different prompting methods, we kept using the same set of demonstration samples and modified their format according to the prompting method.}
To demonstrate the usability of the results reimplemented in our work, we conducted a performance survey on existing baseline prompting with LLMs of comparable strength to those studied in this paper, with the results presented in Appendix~\ref{sec:baseline_result}.

\subsection{Experimental Settings}
\paragraph{Demonstration examples}
Under any prompting method, one dataset is used with the number of demonstration examples in all the experiments discussed in this work.
Specifically, as shown in Tab.~\ref{tab:statistics}, there are 8 demonstration examples each of GSM8K, ASDiv, MArith, and AQUA, 6 examples for StrategyQA, 10 examples for Date, 4 examples for MATH.

\paragraph{Hyperparameters of Greedy Decoding}
We use the vllm library~\footnote{\url{https://github.com/vllm-project/vllm}} for few-shot evaluation. 
For greedy decoding, the hyperparameters are set as: top\_p=1, max\_tokens=500, temperature=0, and the number of reasoning path n=1. 
For self-consistency, the number of reasoning path n is set to 4, 16, 32, 64, 128, and temperature = 0.4. Other hyperparameters are set the same as the greedy decoding.
All inference experiments are based on four A100 GPUs.

\section{Experiments and Results}

\begin{table*}[htb]
 \centering \footnotesize
\renewcommand\arraystretch{0.63}  
    \begin{tabular}{lrlccccccccccc}
    \toprule
    \multicolumn{2}{l}{Method} & \multicolumn{1}{l}{HSP} & G8K   & ASDiv & MArith & AQUA  & SQA   & Date  & \multicolumn{1}{l}{Avg} & \multicolumn{1}{l}{Improvement} &       &       &  \\
    \midrule
    \multirow{9}[2]{*}{Lm2-7B} & \multicolumn{1}{r}{\multirow{2}[1]{*}{SD}} & $\times$ & \textcolor[rgb]{ 0,  .69,  .314}{5.8} & 43.7  & \textcolor[rgb]{ 0,  .69,  .314}{7.4} & 19.7  & 62.0  & 33.1  & 28.6  & \multicolumn{4}{c}{\multirow{9}[2]{*}{\includegraphics[scale=0.38]{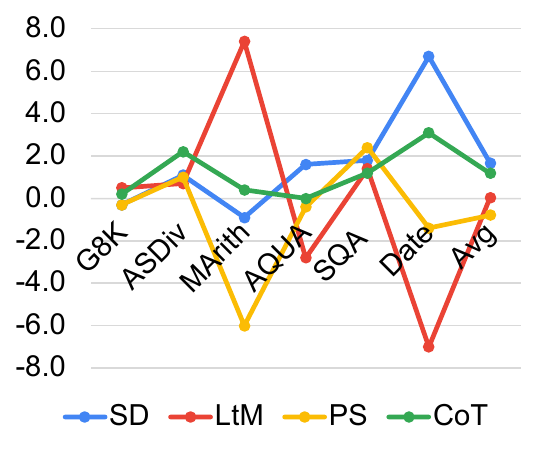}}} \\
          &       & $\checkmark$ $^\dag$ & 5.5   & \textcolor[rgb]{ .898,  .263,  .875}{44.8} & 6.5   & \textcolor[rgb]{ .898,  .263,  .875}{21.3} & \textcolor[rgb]{ .898,  .263,  .875}{63.8} & \textcolor[rgb]{ .898,  .263,  .875}{39.8} & \textcolor[rgb]{ .898,  .263,  .875}{30.3} & \multicolumn{4}{c}{} \\
          & \multicolumn{1}{r}{\multirow{2}[0]{*}{LtM}} & $\times$ & 15.5  & 49.5  & 21.8  & \textcolor[rgb]{ 0,  .69,  .314}{26.0} & 63.9  & \textcolor[rgb]{ 0,  .69,  .314}{49.3} & \textcolor[rgb]{ 0,  .69,  .314}{37.7} & \multicolumn{4}{c}{} \\
          &       & $\checkmark$ $^\dag$ & \textcolor[rgb]{ .957,  0,  .878}{16.0} & \textcolor[rgb]{ .898,  .263,  .875}{50.2} & \textcolor[rgb]{ .898,  .263,  .875}{29.2} & 23.2  & \textcolor[rgb]{ .898,  .263,  .875}{65.3} & 42.3  & \textcolor[rgb]{ .898,  .263,  .875}{37.7} & \multicolumn{4}{c}{} \\
          & \multicolumn{1}{r}{\multirow{2}[0]{*}{PS}} & $\times$ $^\ddag$ & \textcolor[rgb]{ 0,  .663,  .2}{21.8} & 55.8  & \textcolor[rgb]{ 0,  .663,  .2}{66.6} & \textcolor[rgb]{ 0,  .663,  .2}{25.6} & 58.1  & \textcolor[rgb]{ 0,  .69,  .314}{34.8} & 43.8  & \multicolumn{4}{c}{} \\
          &       & $\checkmark$ & 21.5  & \textcolor[rgb]{ .898,  .263,  .875}{56.8} & 60.6  & 25.2  & \textcolor[rgb]{ .957,  0,  .878}{60.5} & 33.4  & \textcolor[rgb]{ .898,  .263,  .875}{43.0} & \multicolumn{4}{c}{} \\
          & \multicolumn{1}{r}{\multirow{2}[0]{*}{CoT}} & $\times$ & 19.7  & 53.6  & 63.4  & \textcolor[rgb]{ 0,  .69,  .314}{24.4} & 66.3  & 40.1  & 44.6  & \multicolumn{4}{c}{} \\
          &       & $\checkmark$ $^\dag$ & \textcolor[rgb]{ .898,  .263,  .875}{19.9} & \textcolor[rgb]{ .898,  .263,  .875}{55.8} & \textcolor[rgb]{ .898,  .263,  .875}{63.8} & \textcolor[rgb]{ .898,  .263,  .875}{24.4} & \textcolor[rgb]{ .898,  .263,  .875}{67.5} & \textcolor[rgb]{ .898,  .263,  .875}{43.2} & \textcolor[rgb]{ .898,  .263,  .875}{45.8} & \multicolumn{4}{c}{} \\
          \cmidrule(lr){2-10}
          &       \multicolumn{2}{c}{ \textcolor[rgb]{ .259,  .522,  .957}{Rlt Avg}} & \textcolor[rgb]{ .259,  .522,  .957}{0.0} & \textcolor[rgb]{ .259,  .522,  .957}{1.2} & \textcolor[rgb]{ .259,  .522,  .957}{0.2} & \textcolor[rgb]{ .259,  .522,  .957}{-0.4} & \textcolor[rgb]{ .259,  .522,  .957}{1.7} & \textcolor[rgb]{ .259,  .522,  .957}{0.3} & \textcolor[rgb]{ .259,  .522,  .957}{0.5} & \multicolumn{4}{c}{} \\
    \midrule
    \multirow{9}[2]{*}{Lm2-13B} & \multicolumn{1}{r}{\multirow{2}[1]{*}{SD}} & $\times$ & \textcolor[rgb]{ 0,  .69,  .314}{8.5} & 48.6  & 10.1  & 19.3  & 65.3  & 40.7  & 32.1  & \multicolumn{4}{c}{\multirow{9}[2]{*}{\includegraphics[scale=0.38]{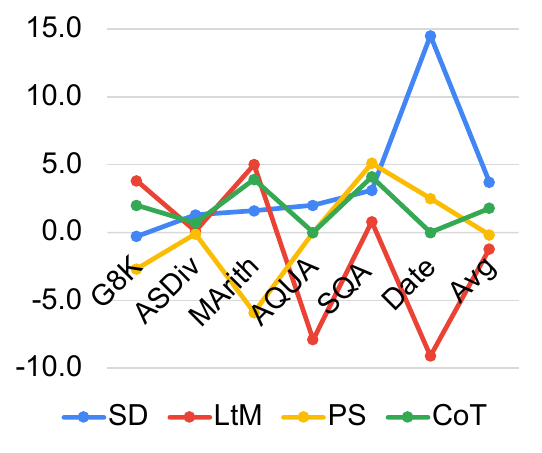}}} \\
          &       & $\checkmark$ $^\dag$ & 8.2   & \textcolor[rgb]{ .898,  .263,  .875}{49.9} & \textcolor[rgb]{ .898,  .263,  .875}{11.7} & \textcolor[rgb]{ .898,  .263,  .875}{21.3} & \textcolor[rgb]{ .898,  .263,  .875}{68.4} & \textcolor[rgb]{ .898,  .263,  .875}{55.2} & \textcolor[rgb]{ .898,  .263,  .875}{35.8} & \multicolumn{4}{c}{} \\
          & \multicolumn{1}{r}{\multirow{2}[0]{*}{LtM}} & $\times$ & 23.8  & 55.8  & 52.7  & \textcolor[rgb]{ 0,  .69,  .314}{31.1} & 68.8  & \textcolor[rgb]{ 0,  .69,  .314}{60.4} & \textcolor[rgb]{ 0,  .69,  .314}{48.8} & \multicolumn{4}{c}{} \\
          &       & $\checkmark$ $^\dag$ & \textcolor[rgb]{ .898,  .263,  .875}{27.6} & \textcolor[rgb]{ .898,  .263,  .875}{55.9} & \textcolor[rgb]{ .898,  .263,  .875}{57.7} & 23.2  & \textcolor[rgb]{ .898,  .263,  .875}{69.6} & 51.3  & 47.6  & \multicolumn{4}{c}{} \\
          & \multicolumn{1}{r}{\multirow{2}[0]{*}{PS}} & $\times$ $^\ddag$ & \textcolor[rgb]{ 0,  .663,  .2}{35.1} & \textcolor[rgb]{ 0,  .663,  .2}{63.0} & \textcolor[rgb]{ 0,  .663,  .2}{80.7} & \textcolor[rgb]{ 0,  .69,  .314}{25.6} & 60.9  & 47.6  & 52.2  & \multicolumn{4}{c}{} \\
          &       & $\checkmark$ & 32.4  & 62.9  & 74.8  & \textcolor[rgb]{ .957,  0,  .878}{25.6} & \textcolor[rgb]{ .957,  0,  .878}{66.0} & \textcolor[rgb]{ .957,  0,  .878}{50.1} & \textcolor[rgb]{ .898,  .263,  .875}{52.0} & \multicolumn{4}{c}{} \\
          & \multicolumn{1}{r}{\multirow{2}[0]{*}{CoT}} & $\times$ & 34.5  & 60.5  & 83.2  & \textcolor[rgb]{ 0,  .69,  .314}{25.6} & 68.0  & \textcolor[rgb]{ 0,  .69,  .314}{57.7} & 54.9  & \multicolumn{4}{c}{} \\
          &       & $\checkmark$ $^\dag$ & \textcolor[rgb]{ .898,  .263,  .875}{36.5} & \textcolor[rgb]{ .898,  .263,  .875}{61.2} & \textcolor[rgb]{ .898,  .263,  .875}{87.1} & \textcolor[rgb]{ .898,  .263,  .875}{25.6} & \textcolor[rgb]{ .898,  .263,  .875}{72.1} & \textcolor[rgb]{ .898,  .263,  .875}{57.7} & \textcolor[rgb]{ .898,  .263,  .875}{56.7} & \multicolumn{4}{c}{} \\
          \cmidrule(lr){2-10}
          &       \multicolumn{2}{c}{ \textcolor[rgb]{ .259,  .522,  .957}{Rlt Avg}} & \textcolor[rgb]{ .259,  .522,  .957}{0.7} & \textcolor[rgb]{ .259,  .522,  .957}{0.5} & \textcolor[rgb]{ .259,  .522,  .957}{1.1} & \textcolor[rgb]{ .259,  .522,  .957}{-1.5} & \textcolor[rgb]{ .259,  .522,  .957}{3.3} & \textcolor[rgb]{ .259,  .522,  .957}{2.0} & \textcolor[rgb]{ .259,  .522,  .957}{1.0} & \multicolumn{4}{c}{} \\
    \midrule
    \multirow{9}[2]{*}{Lm2-70B} & \multicolumn{1}{r}{\multirow{2}[1]{*}{SD}} & $\times$ & 12.6  & 60.6  & \textcolor[rgb]{ 0,  .69,  .314}{26.3} & 24.8  & 72.9  & 54.6  & 42.0  & \multicolumn{4}{c}{\multirow{9}[2]{*}{\includegraphics[scale=0.38]{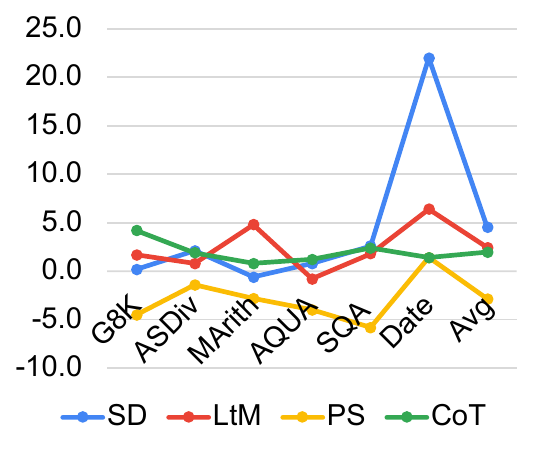}}} \\
          &       & $\checkmark$ $^\dag$ & \textcolor[rgb]{ .898,  .263,  .875}{12.8} & \textcolor[rgb]{ .898,  .263,  .875}{62.7} & 25.7  & \textcolor[rgb]{ .898,  .263,  .875}{25.6} & \textcolor[rgb]{ .898,  .263,  .875}{75.5} & \textcolor[rgb]{ .898,  .263,  .875}{76.6} & \textcolor[rgb]{ .898,  .263,  .875}{46.5} & \multicolumn{4}{c}{} \\
          & \multicolumn{1}{r}{\multirow{2}[0]{*}{LtM}} & $\times$ & 40.2  & 68.6  & 72.0  & \textcolor[rgb]{ 0,  .69,  .314}{39.4} & 75.2  & 71.0  & 61.1  & \multicolumn{4}{c}{} \\
          &       & $\checkmark$ $^\dag$ & \textcolor[rgb]{ .898,  .263,  .875}{41.9} & \textcolor[rgb]{ .898,  .263,  .875}{69.4} & \textcolor[rgb]{ .898,  .263,  .875}{76.8} & 38.6  & \textcolor[rgb]{ .898,  .263,  .875}{77.0} & \textcolor[rgb]{ .898,  .263,  .875}{77.4} & \textcolor[rgb]{ .898,  .263,  .875}{63.5} & \multicolumn{4}{c}{} \\
          & \multicolumn{1}{r}{\multirow{2}[0]{*}{PS}} & $\times$ $^\ddag$ & \textcolor[rgb]{ 0,  .663,  .2}{60.0} & \textcolor[rgb]{ 0,  .663,  .2}{74.1} & \textcolor[rgb]{ 0,  .663,  .2}{95.8} & \textcolor[rgb]{ 0,  .69,  .314}{40.2} & \textcolor[rgb]{ 0,  .69,  .314}{64.7} & 62.4  & 66.2  & \multicolumn{4}{c}{} \\
          &       & $\checkmark$ & 55.5  & 72.7  & 93.0  & 36.2  & 58.9  & \textcolor[rgb]{ .957,  0,  .878}{63.8} & \textcolor[rgb]{ .898,  .263,  .875}{63.4} & \multicolumn{4}{c}{} \\
          & \multicolumn{1}{r}{\multirow{2}[0]{*}{CoT}} & $\times$ & 46.1  & 72.5  & 93.8  & 35.8  & 74.6  & 71.6  & 65.7  & \multicolumn{4}{c}{} \\
          &       & $\checkmark$ $^\dag$ & \textcolor[rgb]{ .898,  .263,  .875}{50.3} & \textcolor[rgb]{ .898,  .263,  .875}{74.4} & \textcolor[rgb]{ .898,  .263,  .875}{94.6} & \textcolor[rgb]{ .898,  .263,  .875}{37.0} & \textcolor[rgb]{ .898,  .263,  .875}{77.0} & \textcolor[rgb]{ .898,  .263,  .875}{73.0} & \textcolor[rgb]{ .898,  .263,  .875}{67.7} & \multicolumn{4}{c}{} \\
          \cmidrule(lr){2-10}
          &       \multicolumn{2}{c}{ \textcolor[rgb]{ .259,  .522,  .957}{Rlt Avg}}& \textcolor[rgb]{ .259,  .522,  .957}{0.4} & \textcolor[rgb]{ .259,  .522,  .957}{0.9} & \textcolor[rgb]{ .259,  .522,  .957}{0.6} & \textcolor[rgb]{ .259,  .522,  .957}{-0.7} & \textcolor[rgb]{ .259,  .522,  .957}{0.2} & \textcolor[rgb]{ .259,  .522,  .957}{7.8} & \textcolor[rgb]{ .259,  .522,  .957}{1.5} & \multicolumn{4}{c}{} \\
    \midrule
    \multirow{9}[2]{*}{Mix-56B} & \multicolumn{1}{r}{\multirow{2}[1]{*}{SD}} & $\times$ & 19.8  & 64.3  & \textcolor[rgb]{ 0,  .69,  .314}{44.6} & 22.0  & \textcolor[rgb]{ 0,  .69,  .314}{72.1} & 45.4  & 44.7  & \multicolumn{4}{c}{\multirow{9}[2]{*}{\includegraphics[scale=0.38]{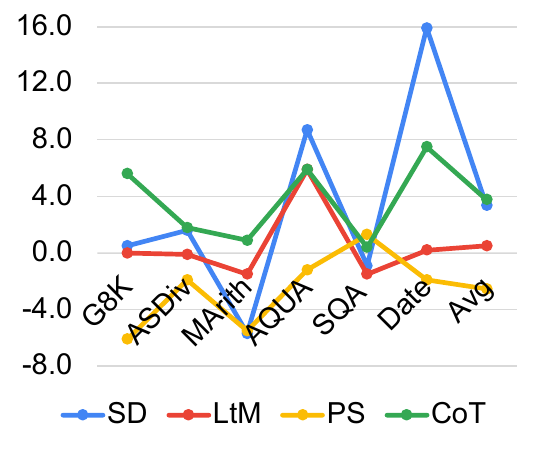}}} \\
          &       & $\checkmark$ $^\dag$ & \textcolor[rgb]{ .898,  .263,  .875}{20.3} & \textcolor[rgb]{ .898,  .263,  .875}{65.9} & 38.9  & \textcolor[rgb]{ .898,  .263,  .875}{30.7} & 71.2  & \textcolor[rgb]{ .898,  .263,  .875}{61.3} & \textcolor[rgb]{ .898,  .263,  .875}{48.1} & \multicolumn{4}{c}{} \\
          & \multicolumn{1}{r}{\multirow{2}[0]{*}{LtM}} & $\times$ $^\ddag$ & \textcolor[rgb]{ 0,  .69,  .314}{56.0} & \textcolor[rgb]{ 0,  .69,  .314}{77.1} & \textcolor[rgb]{ 0,  .69,  .314}{74.3} & 43.3  & \textcolor[rgb]{ 0,  .69,  .314}{73.9} & 64.1  & 64.8  & \multicolumn{4}{c}{} \\
          &       & $\checkmark$ & \textcolor[rgb]{ .898,  .263,  .875}{56.0} & 77.0  & 72.8  & \textcolor[rgb]{ .898,  .263,  .875}{49.2} & 72.4  & \textcolor[rgb]{ .898,  .263,  .875}{64.3} & \textcolor[rgb]{ .898,  .263,  .875}{65.3} & \multicolumn{4}{c}{} \\
          & \multicolumn{1}{r}{\multirow{2}[0]{*}{PS}} & $\times$ $^\ddag$ & \textcolor[rgb]{ 0,  .663,  .2}{73.2} & \textcolor[rgb]{ 0,  .663,  .2}{84.2} & \textcolor[rgb]{ 0,  .663,  .2}{97.8} & \textcolor[rgb]{ 0,  .663,  .2}{49.6} & 66.3  & \textcolor[rgb]{ 0,  .69,  .314}{68.5} & 73.3  & \multicolumn{4}{c}{} \\
          &       & $\checkmark$ & 67.1  & 82.3  & 92.3  & 48.4  & \textcolor[rgb]{ .957,  0,  .878}{67.6} & 66.6  & \textcolor[rgb]{ .898,  .263,  .875}{70.7} & \multicolumn{4}{c}{} \\
          & \multicolumn{1}{r}{\multirow{2}[0]{*}{CoT}} & $\times$ & 63.7  & 78.3  & 96.1  & 42.5  & 74.7  & 69.9  & 70.9  & \multicolumn{4}{c}{} \\
          &       & $\checkmark$ $^\dag$ & \textcolor[rgb]{ .898,  .263,  .875}{69.8} & \textcolor[rgb]{ .898,  .263,  .875}{80.1} & \textcolor[rgb]{ .898,  .263,  .875}{97.0} & \textcolor[rgb]{ .898,  .263,  .875}{48.4} & \textcolor[rgb]{ .898,  .263,  .875}{75.1} & \textcolor[rgb]{ .898,  .263,  .875}{77.4} & \textcolor[rgb]{ .898,  .263,  .875}{74.6} & \multicolumn{4}{c}{} \\
          \cmidrule(lr){2-10}
          &       \multicolumn{2}{c}{ \textcolor[rgb]{ .259,  .522,  .957}{Rlt Avg}}& \textcolor[rgb]{ .259,  .522,  .957}{0.1} & \textcolor[rgb]{ .259,  .522,  .957}{0.4} & \textcolor[rgb]{ .259,  .522,  .957}{-2.9} & \textcolor[rgb]{ .259,  .522,  .957}{4.8} & \textcolor[rgb]{ .259,  .522,  .957}{-0.2} & \textcolor[rgb]{ .259,  .522,  .957}{5.4} & \textcolor[rgb]{ .259,  .522,  .957}{1.3} & \multicolumn{4}{c}{} \\
    \bottomrule
    \end{tabular}%
\caption{Results of applying HSP to existing prompting (Sec.~\ref{sec:baselines}). \textcolor[rgb]{ .204,  .659,  .325}{Green} (\textcolor[rgb]{ .898,  .263,  .875}{pink}) values indicate the best performance without HSP (with HSP). \textit{\textcolor[rgb]{ .259,  .522,  .957}{Rlt Avg}} denotes the average relative improvement on the four prompting methods.
\textit{Improvement} represents the relative performance improvement when introducing HSP compared to not using HSP.  
$^\dag$ indicates HSP significantly boosts performance, whereas $^\ddag$ suggests omitting HSP leads to better results.
      }
  \label{tab:effect_hsp}%
\end{table*}%

\subsection{Q1: Can HSP Work?}
\label{sec:can_work}
In this section, we considered three perspectives to answer whether HSP can enhance LLMs' performance by generating hints containing specific knowledge, pivotal concepts, or analytical insights critical for solving the problem before attempting to solve it.               
Next, we will illustrate the three perspectives in detail.

\subsubsection{Exp-I: When HSP Meets Existing Prompting Methods}
\label{sec:exp1_hsp}
We applied HSP to four existing popular prompting methods to explore how HSP performs in different prompting methods.
Our experimental prompting methods include standard prompting (SD), Least to Most prompting (LtM), Plan-and-Solve prompting (PS), and CoT prompting, as introduced in Sec.~\ref{sec:baselines}
The results are shown in Tab.~\ref{tab:effect_hsp}.
The main findings are summarized as below:

\noindent
(1) \textit{HSP is effective in standard and CoT prompting but fails in PS and LtM prompting.}
From Tab.~\ref{tab:effect_hsp}, we observe that the standard and CoT Prompting show significant performance improvements under HSP, while the enhancements from PS and LtM are limited.
We try to give reasons below: 
Hints clarify the prompt or problem by offering key insights or solutions, influencing the logic behind the answers. They are crucial in task planning for both PS and LtM prompting, where introducing hints early can impact their planning process. Conversely, Standard and CoT prompting, focusing solely on the final answer or intermediate reasoning, are compatible with hints.

\noindent
(2) \textit{Larger model sizes tend to show more significant performance improvements.} 
From Tab.~\ref{tab:main_result}, we can observe that the average performance improvements for 7B, 13B, 56B, and 70B models across four prompting methods (e.g., CoT and LtM) are 0.5, 1.0, 1.3, and 1.5, respectively. 
The reason can be that the model capabilities increase as the size increases, and higher capabilities will help achieve higher quality hints for better problem-solving.

\noindent
(3) \textit{The introduction of HSP can steadily enhance the performance of CoT prompting.}
We observe that CoT, combined with HSP, shows performance enhancements across all four models and six datasets, while SD, LtM, and PS all experience some scenarios of performance drop. 
From the line chart in Tab.~\ref{tab:effect_hsp}, we can observe that LtM and PS exhibit significant fluctuations in average performance gains across each dataset, with numerous settings of negative improvement.

\begin{table*}[htb]
  \centering \footnotesize
  \renewcommand\tabcolsep{2.9pt}
    \begin{tabular}{llccccccccccccccc}
    \toprule
   \multicolumn{2}{c}{Method} & G8K   & ASDiv & MArith & AQUA  & SQA   & Date  & \multicolumn{1}{l}{Avg} & \multicolumn{6}{c}{Improvement} \\
    \midrule
    \multirow{3}[2]{*}{Lm2-7B} & CoT   & 19.7  & 53.6  & 63.4  & 24.4  & 66.3  & 40.1  & 44.6  & \multicolumn{4}{c}{\multirow{6}[4]{*}{\includegraphics[scale=0.33]{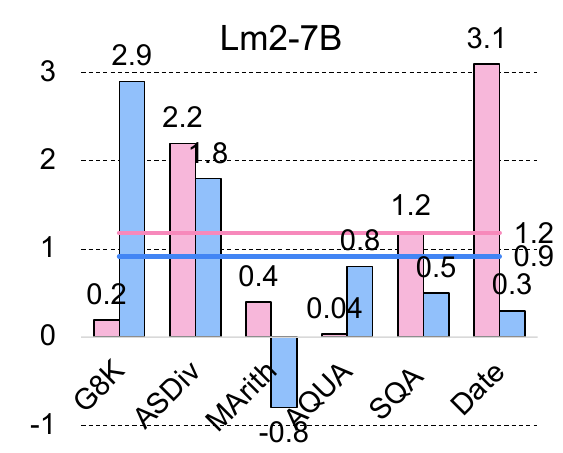}}} & \multicolumn{4}{c}{\multirow{6}[4]{*}{\includegraphics[scale=0.33]{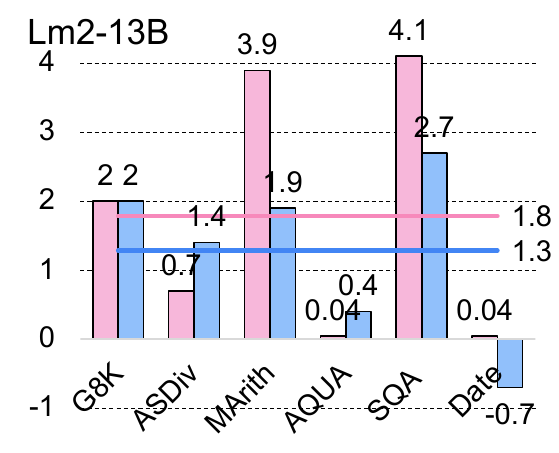}}} \\
          & +HSP$^\dag$  & 19.9  & \textbf{55.8} & \textbf{63.8} & 24.4  & \textbf{67.5} & \textbf{43.2} & \textbf{45.8} & \multicolumn{3}{c}{}  & \multicolumn{3}{c}{} \\
          & +HSP2$^\ddag$ & \textbf{22.6} & 55.4  & 62.6  & \textbf{25.2} & 66.8  & 40.4  & 45.5  & \multicolumn{3}{c}{}  & \multicolumn{3}{c}{} \\
\cmidrule{1-9}    \multirow{3}[2]{*}{Lm2-13B} & CoT   & 34.5  & 60.5  & 83.2  & 25.6  & 68.0  & 57.7  & 54.9  & \multicolumn{3}{c}{}  & \multicolumn{3}{c}{} \\
          & +HSP$^\dag$  & \textbf{36.5} & 61.2  & \textbf{87.1} & 25.6  & \textbf{72.1} & \textbf{57.7} & \textbf{56.7} & \multicolumn{3}{c}{}  & \multicolumn{3}{c}{} \\
          & +HSP2$^\ddag$ & \textbf{36.5} & \textbf{61.9} & 85.1  & \textbf{26.0} & 70.7  & 57.0  & 56.2  & \multicolumn{3}{c}{}  & \multicolumn{3}{c}{} \\
\midrule   \multirow{3}[2]{*}{Lm2-70B} & CoT   & 46.1  & 72.5  & 93.8  & 35.8  & 74.6  & 71.6  & 65.7  & \multicolumn{4}{c}{\multirow{6}[4]{*}{\includegraphics[scale=0.33]{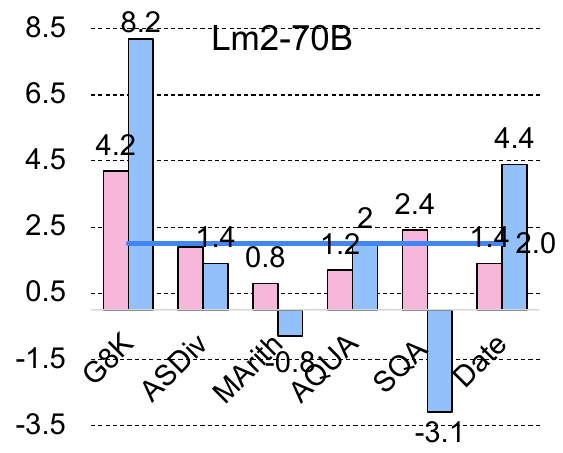}}} & \multicolumn{4}{c}{\multirow{6}[4]{*}{\includegraphics[scale=0.33]{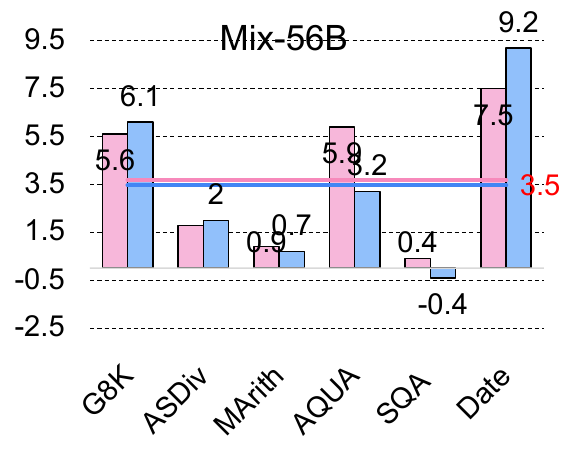}}} \\
          & +HSP$^\dag$  & 50.3  & \textbf{74.4} & \textbf{94.6} & 37.0  & \textbf{77.0} & 73.0  & 67.7  & \multicolumn{3}{c}{}  & \multicolumn{3}{c}{} \\
          & +HSP2$^\ddag$ & \textbf{54.3} & 73.9  & 93.0  & \textbf{37.8} & 71.5  & \textbf{76.0} & \textbf{67.8} & \multicolumn{3}{c}{}  & \multicolumn{3}{c}{} \\
\cmidrule{1-9}    \multirow{3}[2]{*}{Mix-56B} & CoT   & 63.7  & 78.3  & 96.1  & 42.5  & 74.7  & 69.9  & 70.9  & \multicolumn{3}{c}{}  & \multicolumn{3}{c}{} \\
          & +HSP$^\dag$  & 69.3  & 80.1  & \textbf{97.0} & \textbf{48.4} & \textbf{75.1} & 77.4  & \textbf{74.6} & \multicolumn{3}{c}{}  & \multicolumn{3}{c}{} \\
          & +HSP2$^\ddag$ & \textbf{69.8} & \textbf{80.3} & 96.8  & 45.7  & 74.3  & \textbf{79.1} & 74.3  & \multicolumn{3}{c}{}  & \multicolumn{3}{c}{} \\
\bottomrule
\end{tabular}%

  \caption{The results of applying HSP and HSP2 in CoT prompting. The \textbf{bold} values indicate the best performance. $^\dag$ and $^\ddag$ denote that the performance of HSP and HSP2 is significantly better than CoT prompting, respectively.}
  \label{tab:main_result}%
\end{table*}%

\begin{table}[htbp]
    \centering \footnotesize
    \renewcommand\arraystretch{0.8}  
  \renewcommand\tabcolsep{1.6pt}
    \begin{tabular}{ccccccccc}
    \toprule
    \multicolumn{2}{c}{Method} & G8K   & ASDiv & MArith & AQUA  & SQA   & Date  & \multicolumn{1}{l}{Avg} \\
    \midrule
    \multicolumn{2}{l}{ChatGPT} & 79.1  & -    & 97.3  & 55.1  & -    & -    & - \\
    \midrule
    \multicolumn{1}{l}{7B} & \multicolumn{1}{l}{HSP2} & 22.6  & 55.4  & 62.6  & 25.2  & 66.8  & 40.4  & 45.5 \\
          & \multicolumn{1}{l}{HSP2G} & \textbf{39.0} & \textbf{62.5} & \textbf{88.9} & \textbf{28.7} & \textbf{69.5} & \textbf{61.0} & \textbf{58.3} \\
          & \multicolumn{1}{l}{\textcolor[rgb]{ 0,  .69,  .314}{Impv}} & \textcolor[rgb]{ 0,  .69,  .314}{16.4} & \textcolor[rgb]{ 0,  .69,  .314}{7.1} & \textcolor[rgb]{ 0,  .69,  .314}{26.3} & \textcolor[rgb]{ 0,  .69,  .314}{3.5} & \textcolor[rgb]{ 0,  .69,  .314}{2.7} & \textcolor[rgb]{ 0,  .69,  .314}{20.6} & \textcolor[rgb]{ 0,  .69,  .314}{12.8} \\
    \midrule
    \multicolumn{1}{l}{13B} & \multicolumn{1}{l}{HSP2} & 36.5  & 61.9  & 85.1  & 26.0  & 70.7  & 57.0  & 56.2 \\
          & \multicolumn{1}{l}{HSP2G} & \textbf{56.4} & \textbf{66.4} & \textbf{95.6} & \textbf{36.6} & \textbf{72.0} & \textbf{69.4} & \textbf{66.1} \\
          & \multicolumn{1}{l}{\textcolor[rgb]{ 0,  .69,  .314}{Impv}} & \textcolor[rgb]{ 0,  .69,  .314}{19.9} & \textcolor[rgb]{ 0,  .69,  .314}{4.5} & \textcolor[rgb]{ 0,  .69,  .314}{10.5} & \textcolor[rgb]{ 0,  .69,  .314}{10.6} & \textcolor[rgb]{ 0,  .69,  .314}{1.3} & \textcolor[rgb]{ 0,  .69,  .314}{12.4} & \textcolor[rgb]{ 0,  .69,  .314}{9.9} \\
    \midrule
    \multicolumn{1}{l}{70B} & \multicolumn{1}{l}{HSP2} & 54.3  & 73.9  & 93.0  & 37.8  & 71.5  & 76.0  & 67.8 \\
          & \multicolumn{1}{l}{HSP2G} & \textbf{68.2} & \textbf{79.0} & \textbf{98.0} & \textbf{43.3} & \textbf{76.6} & \textbf{87.7} & \textbf{75.5} \\
          & \multicolumn{1}{l}{\textcolor[rgb]{ 0,  .69,  .314}{Impv}} & \textcolor[rgb]{ 0,  .69,  .314}{13.9} & \textcolor[rgb]{ 0,  .69,  .314}{5.1} & \textcolor[rgb]{ 0,  .69,  .314}{5.0} & \textcolor[rgb]{ 0,  .69,  .314}{5.5} & \textcolor[rgb]{ 0,  .69,  .314}{5.1} & \textcolor[rgb]{ 0,  .69,  .314}{11.7} & \textcolor[rgb]{ 0,  .69,  .314}{7.7} \\
    \midrule
    \multicolumn{1}{l}{56B} & \multicolumn{1}{l}{HSP2} & 69.8  & 80.3  & 96.8  & 45.7  & 74.3  & 79.1  & 74.3 \\
          & \multicolumn{1}{l}{HSP2G} & \textbf{79.5} & \textbf{84.1} & \textbf{99.2} & \textbf{56.3} & \textbf{76.5} & \textbf{84.7} & \textbf{80.1} \\
          & \multicolumn{1}{l}{\textcolor[rgb]{ 0,  .69,  .314}{Impv}} & \textcolor[rgb]{ 0,  .69,  .314}{9.7 } & \textcolor[rgb]{ 0,  .69,  .314}{3.8 } & \textcolor[rgb]{ 0,  .69,  .314}{2.4 } & \textcolor[rgb]{ 0,  .69,  .314}{10.6 } & \textcolor[rgb]{ 0,  .69,  .314}{2.2 } & \textcolor[rgb]{ 0,  .69,  .314}{5.6 } & \textcolor[rgb]{ 0,  .69,  .314}{5.7 } \\
    \midrule
    \multicolumn{2}{c}{\textcolor[rgb]{ .373,  .616,  .984}{Avg impv}} & \textcolor[rgb]{ .373,  .616,  .984}{15.0} & \textcolor[rgb]{ .373,  .616,  .984}{5.1} & \textcolor[rgb]{ .373,  .616,  .984}{11.1} & \textcolor[rgb]{ .373,  .616,  .984}{7.6} & \textcolor[rgb]{ .373,  .616,  .984}{2.8} & \textcolor[rgb]{ .373,  .616,  .984}{12.6} & \textcolor[rgb]{ .373,  .616,  .984}{9.0} \\
    \bottomrule
    \end{tabular}%
    \vspace{-7pt}
\caption{Experimental results of enhancing HSP2 with hints generated by GPT4. The  values in green are the performance gap between HSP2G and HSP2. The blue values are the  improvement across the four models.
The values in bold represent the best performance. } 
  \label{tab:gpt4_hint}%
\end{table}%

\subsubsection{EXP-II: Effectiveness of HSP for CoT Prompting}
\label{sec:comp_variants}
In Exp-I, we found that applying HSP to CoT prompting results in significant and stable performance improvements across six datasets.
Based on this, to identify flexible and effective ways to incorporate HSP, we attempted to explore whether a two-stage HSP (HSP2) approach could work in CoT prompting.
The two-stage HSP means that LLMs produce outputs twice, first outputting a hint and then a solution. In contrast, HSP has only one output that contains both the hint and the solution.
Experimental results on 6 datasets of 4 open source models are shown in Tab.~\ref{tab:main_result}.
The main observations are summarized as below:

\noindent
(1) \textit{The performance of HSP and HSP2 is comparable, despite the different ways of introducing hints.}
We can observe that among four LLMs, the largest average performance gap between HSP and HSP2 across six datasets was achieved on the Llama2-13B model with 0.5\% (56.7-56.2).
This indicates that although the methods of introducing hints differ, the extent of performance improvement brought by both is  close.

\noindent
(2) \textit{HSP brings more stable improvements compared to HSP2.} From histograms in Tab.~\ref{tab:main_result}, HSP shows improvements on nearly every dataset under models of four different sizes. In contrast, HSP2 may lead to performance decreases in certain scenarios, for example, on the MArith dataset, the HSP2 performance decreases with Llama2-7B and Llama2-70B models.

\subsubsection{Exp-III: The Impact of Hint Quality}
\label{sec:high-quality}

Introducing HSP can effectively enhance the performance of CoT prompting. But what is the upper bound?
Here, we choose to explore on HSP2 because it enables the hints from external sources, a feature not available in the one-stage HSP structure, and HSP2 is comparable in strength to HSP (Sec.~\ref{sec:comp_variants}). Hints generated by GPT-4 will be used as part of the input in the HSP2, denoting as HSP2G.
Experimental results are shown in Tab.~\ref{tab:gpt4_hint}. The performance of ChatGPT is copied from~\citet{yin:eot}, where the number of examples used to evaluate GSM8K, MultiArith, and AQUA is 8, 8, and 4, respectively. The main findings are summarized as below:

(1) \textit{High-quality hints make the open-source model outperforms ChatGPT.}
We can observe that with the introduction of high-quality hints, all of the four LLMs with different model sizes and structures  consistently showed performance improvement across six datasets. Furthermore, the Mix-56B equipped with HSP2(GPT4) outperformed ChatGPT on the GSM8K, MultiArith, and AQUA datasets.

(2) \textit{The introduction of high-quality hints leads to more improvements in lower-capability models.}
Tab.~\ref{tab:gpt4_hint} shows that the average performance improvements for the Llama2 models sized 7B, 13B, and 70B are 12.8, 9.9, and 7.7, respectively. 
This indicates that with the support of high-quality hints, HSP2(GPT4)'s performance has improved a lot compared to HSP2.
This can be attributed to that the low capability LLMs are hard to generate helpful hints that can assist in providing correct solutions. 
By providing high-quality hints, it is possible to offer more benefits beyond the capability of lower-ability LLMs. Therefore, there is a relatively large improvement in performance.

\subsection{Q2: Can HSP Work on Hard Tasks?}
\label{sec:hard_task}

\subsubsection{EXP-IV: Exploring Difficult Tasks}
\label{sec:exp4_hard_task}

\begin{table}[htb]
  \centering \footnotesize
   \renewcommand\tabcolsep{2pt}
    \begin{tabular}{lcccc}
    \toprule
    Prompting & Lm2-7B & Lm2-13B & Lm2-70B & Mix-56B \\
    \midrule
    CoT   & \textbf{4.5} & 5.6   & 11.1  & 27.0 \\
    +HSP & 4.4   & \textbf{5.7} & \textbf{11.4} & \textbf{28.6}$^\dag$ \\
    \bottomrule
    \end{tabular}%
      \caption{Results on MATH dataset. Values in bold denote the best performance, and the value with $^\dag$ denotes the performance of HSP significantly outperforms CoT.
      }
  \label{tab:math_eval}%
\end{table}%

\begin{table*}[htb]
  \centering \footnotesize
      \renewcommand\arraystretch{0.89}  
  \renewcommand\tabcolsep{2.8pt}
    \begin{tabular}{llccccccccccccc}
    \toprule
    \multicolumn{1}{l}{\multirow{2}[2]{*}{Param.}} & \multicolumn{1}{c}{\multirow{2}[2]{*}{Prompting}} & \multirow{2}[2]{*}{Overall} & \multicolumn{7}{c}{Type}                              & \multicolumn{5}{c}{Level} \\
    \cmidrule(lr){4-10} \cmidrule(lr){11-15}
          &       &       & AG    & CP    & GT    & IA    & NT    & PG    & PC    & L1    & L2    & L3    & L4    & L5 \\
    \midrule
    \multicolumn{1}{l}{\multirow{3}[1]{*}{n=1,t=0}} & CoT   & 27.0  & 39.01 & 18.99 & 18.58 & 13.4  & 16.85 & 47.07 & 15.57 & 62.47 & 44.41 & 30.59 & 18.62 & 8.08 \\
          & +HSP & 28.6  & 39.09 & 23.21 & 21.09 & 13.84 & 15.93 & 46.27 & 15.2  & 64.3  & 45.64 & 30.33 & 18.29 & 8.91 \\
          & Impv  & 1.62  & \textcolor[rgb]{ 0,  .69,  .314}{0.08} & \textcolor[rgb]{ 0,  .69,  .314}{\textbf{4.22}} & \textcolor[rgb]{ 0,  .69,  .314}{\textbf{2.51}} & \textcolor[rgb]{ 0,  .69,  .314}{0.44} & \textcolor[rgb]{ 1,  0,  0}{-0.92} & \textcolor[rgb]{ 1,  0,  0}{-0.8} & \textcolor[rgb]{ 1,  0,  0}{-0.37} & \textcolor[rgb]{ 0,  .69,  .314}{\textbf{1.83}} & \textcolor[rgb]{ 0,  .69,  .314}{\textbf{1.23}} & \textcolor[rgb]{ 1,  0,  0}{-0.26} & \textcolor[rgb]{ 1,  0,  0}{-0.33} & \textcolor[rgb]{ 0,  .69,  .314}{0.83} \\
    \midrule
    \multicolumn{1}{l}{\multirow{3}[0]{*}{n=4,t=0.4}} & CoT   & 31.9  & 46.67 & 26.58 & 22.55 & 15.39 & 20.56 & 52.47 & 17.95 & 71.17 & 49.33 & 36.6  & 23.39 & 10.8 \\
          & +HSP & 33    & 47.35 & 26.37 & 26.1  & 15.39 & 21.3  & 54.88 & 19.6  & 72.31 & 51.45 & 36.34 & 25.86 & 11.33 \\
          & Impv  & 1.1   & \textcolor[rgb]{ 0,  .69,  .314}{0.68} & \textcolor[rgb]{ 1,  0,  0}{-0.21} & \textcolor[rgb]{ 0,  .69,  .314}{3.55} & \textcolor[rgb]{ 0,  .69,  .314}{0} & \textcolor[rgb]{ 0,  .69,  .314}{0.74} & \textcolor[rgb]{ 0,  .69,  .314}{\textbf{2.41}} & \textcolor[rgb]{ 0,  .69,  .314}{\textbf{1.65}} & \textcolor[rgb]{ 0,  .69,  .314}{\textbf{1.14}} & \textcolor[rgb]{ 0,  .69,  .314}{\textbf{2.12}} & \textcolor[rgb]{ 1,  0,  0}{-0.26} & \textcolor[rgb]{ 0,  .69,  .314}{\textbf{2.47}} & \textcolor[rgb]{ 0,  .69,  .314}{0.53} \\
    \midrule
    \multicolumn{1}{l}{\multirow{3}[1]{*}{n=16,t=0.4}} & CoT   & 37.6  & 53.41 & 31.22 & 27.35 & 19.38 & 26.67 & 58.9  & 24.73 & 78.03 & 56.71 & 43.15 & 30.07 & 13.52 \\
          & +HSP & 38.8  & 53.75 & 32.07 & 31.52 & 20.93 & 27.59 & 59.82 & 26.01 & 78.49 & 57.83 & 44.39 & 33.11 & 13.44 \\
          & Impv  & 1.2   & \textcolor[rgb]{ 0,  .69,  .314}{0.34} & \textcolor[rgb]{ 0,  .69,  .314}{0.85} & \textcolor[rgb]{ 0,  .69,  .314}{\textbf{4.17}} & \textcolor[rgb]{ 0,  .69,  .314}{\textbf{1.55}} & \textcolor[rgb]{ 0,  .69,  .314}{0.92} & \textcolor[rgb]{ 0,  .69,  .314}{0.92} & \textcolor[rgb]{ 0,  .69,  .314}{\textbf{1.28}} & \textcolor[rgb]{ 0,  .69,  .314}{0.46} & \textcolor[rgb]{ 0,  .69,  .314}{\textbf{1.12}} & \textcolor[rgb]{ 0,  .69,  .314}{\textbf{1.24}} & \textcolor[rgb]{ 0,  .69,  .314}{\textbf{3.04}} & \textcolor[rgb]{ 1,  0,  0}{-0.08} \\
    \bottomrule
    \end{tabular}%
      \caption{The results of fine-grained evaluation for Mix-56B on the MATH dataset based on topic and problem difficulty. n is the number of sample paths of the self-consistency, and t is the temperature.
      \textit{AG, CP, GT, IA, NT, PA, PC} respectively represent \textit{Algebra, Counting \& Probability, Geometry, Intermediate Algebra, Number Theory, Prealgebra, Precalculus}.
      \textcolor{green}{Green} values indicate an performance improvement of HSP prompting relative to CoT prompting, while \textcolor{red}{red} values indicate a decrease. Values in bold denote performance improvements greater than 1.}
  \label{tab:math_fine-grained}%
\end{table*}%

The tasks we have explored are those that LLMs can handle well. 
As the difficulty of the task increases, LLMs may not possess sufficient knowledge and capability to address it. 
This raises a research question: Can LLMs generate helpful hints when they meet the challenge task?

To answer this question, we chose to investigate the MATH dataset~\cite{math:hendrycks}, which is a dataset that poses challenges for large language models.
The results are shown in Tab.~\ref{tab:math_eval}. 
We can observe that only the Mix-56B model shows a significant improvement of 1.6 under CoT+HSP prompting, while the Llama-2 family model fails. 
The reason might be that the Llama-2 family models face significant challenges on the MATH dataset, with their best result being only 11.4 (Lm2-70B), while the Mix-56B model achieves 27.0 under CoT prompting, it is difficult for Llama-2 family model to generate valuable hints.

To find which kind of samples Mix-56B can work, we performed a fine-grained analysis based on the mathematic problem topic and the difficulty, where the dataset provides the topics and the difficulty levels.
Furthermore, to explore how self-consistency affects the performance, we evaluate this model using sample paths of n=4 and n=16 and a model temperature of 0.4.
The results are shown in Tab.~\ref{tab:math_fine-grained}.
The main findings can be summarized as:
(1) As n increases, under the CoT+HSP setting, the samples for which the LLM sees performance improvements shift from low to high difficulty. 
(2) As n increases, it is commonly believed that the most challenging GT type experiences the most significant performance improvement, amounting to 4.17.
These indicate that by increasing n, HSP enhancement will correctly solve more complex questions.

\begin{figure}[ht]
	\centering
	\subfigure[7B]{
	\begin{minipage}[b]{0.162 \textwidth}
		\includegraphics[width=1\textwidth]{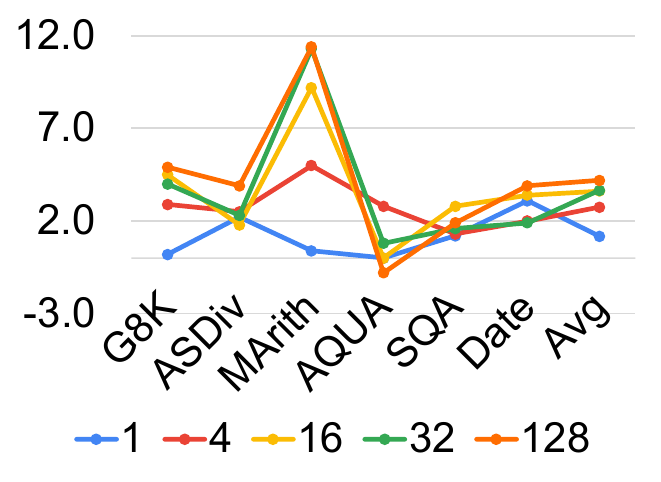}
	\end{minipage}
	} 
	\hspace{-15pt}
 \subfigure[13B]{
	\begin{minipage}[b]{0.162 \textwidth}
		\includegraphics[width=1\textwidth]{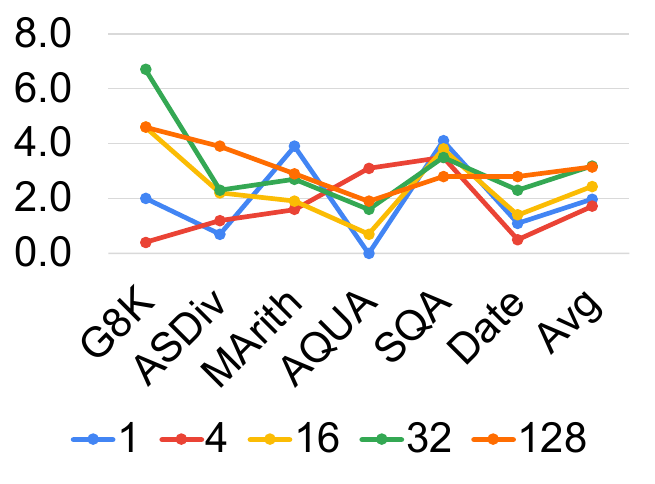}
	\end{minipage}
	}  
        \hspace{-15pt}
	\subfigure[70B]{
	\begin{minipage}[b]{0.162 \textwidth}
		\includegraphics[width=1\textwidth]{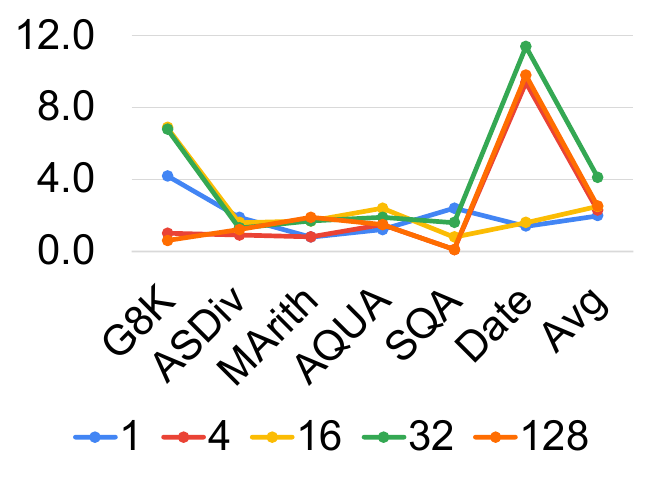}
	\end{minipage}
	}  
 \vspace{-15pt}
	\caption{The relative performance improvement of self-consistency between CoT+HSP and CoT. The numbers of sample paths are 4, 16, 32, and 128, and the model temperature is 0.4.
}
\label{fig:sc_rlt_impv}
\end{figure}

\subsubsection{EXP-V: The Impact of Self-consistency}
In EXP-IV (Sec.~\ref{sec:exp4_hard_task}), we found that self-consistency setting can  improve performance of difficult tasks (MATH dataset), even difficult samples. 
This raises the question of how CoT prompting equipped with HSP performs under a self-consistency setting for the popular tasks.
We sample paths with numbers (n) 4, 16, 32, and 128 for the self-consistency study and set the model temperature as 0.4. The relative improvement between CoT+HSP and CoT on six datasets is shown in Fig.~\ref{fig:sc_rlt_impv} (Full results can be seen in the Appendix~\ref{sec:sc_full_appendix}).
The main findings are as below:

\noindent
(1) \textit{As the number of sampling paths increases, the relative improvements brought by applying HSP also increase.} From Fig.~\ref{fig:sc_rlt_impv}, we can observe that at n=32 or n=128, all three models achieve their best performance.
By calculating the Pearson correlation between the number of sampling  (n) and relative performance for Lm2-7B, Lm2-13B, and Lm2-70B (excluding n=128), the correlations are 0.67, 0.72, and 0.95, respectively. 
The reason can be that the larger n leads to more explored hints, making it easier to generate hints beneficial for problem-solving.

\noindent
(2) \textit{Smaller models see the most significant relative performance improvement after applying self-consistency.} This might be because smaller models have lower capabilities, while with guided hints, increasing n makes it easier to correct originally incorrect solutions, thus leading to more substantial performance improvements.

\subsection{Q3 (EXP-VI): How does SFT Perform on HSP Format Datasets?}
\label{sec:sft}

Despite the remarkable success of LLMs, most existing open-source LLMs (e.g., LLaMA-2) still face challenges in solving math problems due to complex reasoning processes.
How do LLMs perform when they are supervised fine-tuning (SFT) on the HSP format dataset?

We construct a SFT dastaset with CoT+HSP format. Specifically, we collected hints by GPT4 for the GSM8K training set with 7.5k samples. 75,000 samples that rewrite the original questions from the MetaMATH~\cite{metamath:yu}, are extracted. And the hints will be utilized to the derived questions. The dataset is named HSPMATH, and the original 7.5k samples will be used as standard samples, which we call HSPMATH-1.s
The results with supervised fine-tuning on GSM8K under Llemma-7B and Llama2-13B are shown in Tab.~\ref{tab:math_sft}. 
The baselines include: Llama2~\cite{touvron2023llama}, 
RFT~\cite{RFT}, 
Llemma~\cite{Llemma}, 
WizardMath~\cite{WizardMath}, 
WizardLM~\cite{xu2023wizardlm},
MetaMath~\cite{metamath:yu},
GPT-3.5~\cite{openai:gpt4},
PaLM~\cite{palm},
Minerva~\cite{Minerva}, and
Chinchilla~\cite{Chinchilla}
We can observed:

\noindent
(1) \textit{Supervised fine-tuning on datasets with HSP allows LLMs to achieve significant performance improvements.} 
From Tab.~\ref{tab:math_sft}, we can observe that in three groups of SFT (e.g., HSP-Llemma vs. Llemma on the HSPMATH1 dataset), the performance dramatically improves with HSP enhancement, which is 5.1, 12.3, and 5.6, respectively. The reason can be that supervised fine-tuning involving hints helps the model better utilize encoded knowledge during the reasoning stage, thereby improving the model's generalization ability.

(2) \textit{The result of HSP-Llemma-7B surpassed many popular LLMs, including GPT-3.5 and WizardMath.}
By fine-tuning the HSPMATH dataset with 75k CoT+HSP format samples, our HSP-Llemma-7B achieved a competitive performance of 64.3, surpassing closed-source models such as GPT-3.5 (57.1) and PaLM-540B (56.5), and WizardMath-13B (63.9), which was fine-tuned on a large-scale mathematical corpus. It approaches the performance of MetaMath-7B (66.5), fine-tuned on a corpus of 40k samples.

\begin{table}[htbp]
  \centering \footnotesize
  \renewcommand\tabcolsep{2.3pt}
    \begin{tabular}{lcclcc}
    \toprule
    Model  & Size  & ACC   & Model & Size  & ACC \\
    \cmidrule(lr){1-3}\cmidrule(lr){4-6}
    \multicolumn{3}{l}{\textit{open source}} & \multicolumn{3}{l}{\textit{close source}} \\
    \cmidrule(lr){1-3}\cmidrule(lr){4-6}
    Llama2 & 7B    & 14.6  & GPT-3.5 & -     & 57.1 \\
    Llama2 & 13B   & 28.7  & PaLM  & 540B  & 56.5 \\
    Llemma & 7B    & 36.4  & Minerva & 540B  & 58.8 \\
    Llama2 & 34B   & 42.2  & Minerva & 62B   & 52.4 \\
    RFT   & 7B    & 50.3  & Chinchilla & 70B   & 43.7 \\
\cmidrule{4-6}    Llemma & 34B   & 51.5  & \multicolumn{3}{l}{\textit{HSPMATH-1 (7.5k samples)}} \\
\cmidrule{4-6}    RFT   & 13B   & 54.8  & Llemma & 7B    & 46.8 \\
    WizardMath & 7B    & 54.9  & HSP-Llemma & 7B    & 51.9 \\
    WizardLM & 13B   & 55.3  & Llama2 & 13B   & 42.6 \\
    Llama2 & 70B   & 56.8  & HSP-Llama2 & 13B   & 54.9 \\
\cmidrule{4-6}    WizardMath & 13B   & 63.9  & \multicolumn{3}{l}{\textit{HSPMATH (75k samples)}} \\
\cmidrule{4-6}    MetaMath & 7B    & 66.5  & Llemma & 7B    & 58.7 \\
    MetaMath & 13B   & 72.3  & HSP-Llemma & 7B    & \textbf{64.3} \\
    \bottomrule
    \end{tabular}%
    \caption{The results of supervised fine-tuning on GSM8K. The value in bold denotes best SFT result.}
  \label{tab:math_sft}%
\end{table}%

\section{Related Work}
Chain-of-thought (CoT) has given a lot of inspiration to many works and has made numerous attempts to explore high performance.
These techniques include using programming languages to represent the reasoning process~\cite{pal:gao,faithfulcot:lyu}, representing the reasoning process with complex structures such as trees or graphs~\cite{tree:yao,graph:besta}, task decomposition~\cite{zhou2022least,decom1} and combining different prompting~\cite{verify:liu,JianpengZhou}.

For the use of hint enhancement, \citet{php} proposed Progressive-Hint Prompting (PHP), which aims to enhance LLMs' effectiveness by introducing hints iterative, where the hint is a numerical value obtained from the previous solution (or base prompt's solution).
However, the hints for our HSP come from LLMs themselves, while PHP comes from previous predictions. Moreover, our hints can be one-stage, whereas PHP must be multi-staged.

\section{Analysis}

\subsection{Length of Reasoning}
\label{sec:appendix_lengthOfReasoing}
\begin{figure}[!th]
    \centering
    \includegraphics[width=0.92\linewidth]{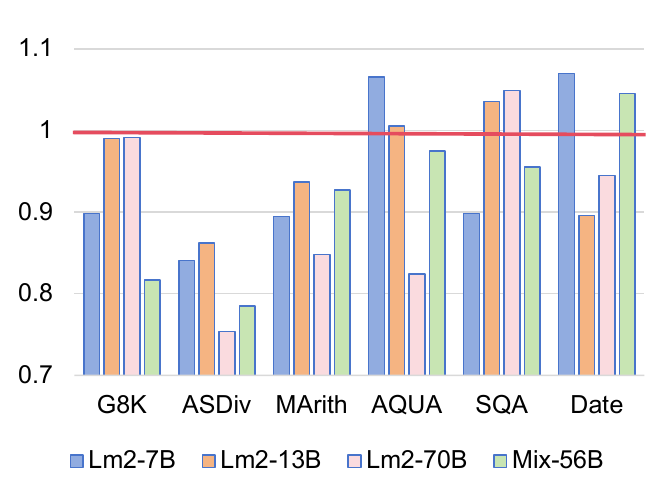}
    \vspace{-8pt}
    \caption{The ratio of solution lengths between CoT and HSP+CoT (HSP applied to CoT prompting). The red line (y=1) indicates that the solution lengths of CoT equals to HSP+CoT.}
	\label{fig:hsp_length_comp}
\end{figure}

Can HSP enhance the model's reasoning capability and effectively reduce the length of the solution generated?
To answer this question, we calculated the solution lengths for CoT and CoT+HSP (applying HSP to CoT). 
For easy understanding, we divided the solution length of CoT+HSP by the solution length of CoT, with the results shown in Fig.~\ref{fig:hsp_length_comp}, where the red horizontal line indicates that the solution lengths of CoT and CoT+HSP are equal.

Our main observation are summarized as below:

\noindent
(1) \textit{Introducing HSP can effectively reduce the length of the solution.} From Fig.~\ref{fig:hsp_length_comp}, we can observate that, out of 24 results across four models and six datasets, only 5 instances show CoT+HSP having a longer solution length than CoT.

\noindent
(2) \textit{The effect of reducing the solution length by introducing HSP is most pronounced in mathematical reasoning tasks.}

\subsection{Case Study}
Guiding the model to generate hints before the solution can effectively improve the model's performance. So, how does guiding LLM to generate hints first affect the generation of the model's solution?
We choose to introduce hints under CoT prompting and select case studies on mathematical reasoning and common sense reasoning tasks, as shown in Tab ~\ref{tab:case_study_part}.

\paragraph{Case 1}
For the question, "Could a Jujutsu expert hypothetically defeat a Janissary?". Under CoT prompting, the LLM-generated solution only explained what ``Jujutsu expert'' and ``Janissary'' are. 
However, in CoT+HSP, the generated hint mentioned analyzing the possibility of the Jujutsu expert defeating Janissary from the perspectives of ``martial arts skills'' and ``weapons,'' thus making a correct solution followed up after introducing the hint.

\paragraph{Case 2}
The solution from CoT seems reasonable, but when calculating the annual total income of a teacher and coach, it was not multiplied by the hourly wage, leading to a final miscalculation. In contrast, CoT+HSP, within the hint, provided the problem-solving ideas, allowing for the correct answer to be calculated step by step in the solution based on the problem-solving strategy mentioned in the hint.

\begin{table*}[!htb]
  \centering \footnotesize
    \begin{tabular}{p{40em}}
    \toprule
    \textbf{Case 1: Commonsense reasoning} \\
    \midrule
    \textcolor{blue}{Question}: Could a Jujutsu expert hypothetically defeat a Janissary? \\
    \textcolor{blue}{True Answer}: FALSE \\
    \textcolor{blue}{CoT Response}: Jujutsu is a martial art that focuses on using an opponent's strength against them.Janissaries were elite soldiers in the Ottoman Empire.Thus, a Jujutsu expert could potentially defeat a Janissary. So the answer is yes. \textcolor{cyan}{[WRONG]} \\
    \textcolor{blue}{CoT+HSP Response}: \\Hint: Compare the \textcolor{redpink}{martial arts skills of a Jujutsu expert with the combat training of a Janissary, considering the historical context and weaponry.} \\Solution: A Jujutsu expert has martial arts skills, but no weapons.A Janissary is an elite Ottoman soldier with advanced combat training and weaponry. Thus, a Janissary would likely defeat a Jujutsu expert. So the answer is no. \textcolor{cyan}{[CORRECT]}\\
    \midrule
    \textbf{Case 2: Mathematical reasoning} \\
    \midrule
    \textcolor{blue}{Question}: Jill gets paid \$20 per hour to teach and \$30 to be a cheerleading coach. If she works 50 weeks a year, 35 hours a week as a teacher and 15 hours a week as a coach, what's her annual salary? \\
    \textcolor{blue}{True Answer}: 57500 \\
   \textcolor{blue}{CoT Response}: Jill works 50 weeks a year. She works 35 hours a week as a teacher, so that's 35 x 50 = \$1750 per year. She works 15 hours a week as a coach, so that's 15 x 50 = \$750 per year. Her total annual salary is \$1750 + \$750 = \$2500. The answer is \$2500. \textcolor{cyan}{[WRONG]}\\
    \textcolor{blue}{CoT+HSP Response}: \\Hint: To find out Jill's annual salary, \textcolor{redpink}{first calculate how much she makes per week as a teacher and as a coach. Then, multiply this total amount by the number of weeks she works in a year. This will give you Jill's annual salary.} \\Solution: Jill makes \$20 per hour as a teacher, so 35 hours a week is 35 x 20 = 700 per week. As a coach, she makes \$30 per hour, so 15 hours a week is 15 x 30 = 450 per week. 700 + 450 = 1150 per week. 1150 x 50 weeks = 57500 per year. The answer is 57500. \textcolor{cyan}{[CORRECT]}\\
    \bottomrule
    \end{tabular}%
    \vspace{-6pt}
      \caption{Case studies of solving mathematical reasoning and common sense reasoning problems with CoT+HSP and CoT prompting on the Mixtral-7*8B model. \textcolor{blue}{Blue} text indicates the stem, \textcolor{redpink}{pink} text indicates the effective hint, \textcolor{cyan}{cyan} text indicates the judgment of whether the answer is correct, \textcolor{cyan}{[CORRECT]} denotes correct, and \textcolor{cyan}{[WRONG]} denotes incorrect.}
  \label{tab:case_study_part}%
\end{table*}%

\section{Conclusion}
In this work, we present Hint-before-Solving Prompting (HSP), a technique that directs Large Language Models (LLMs) to initially produce hints that assist in problem-solving before generating solutions that incorporate intermediate reasoning steps. 
This method alleviates the problem that LLMs, despite having vast knowledge, still encounter in effectively utilizing their encoded knowledge to construct precise and rational reasoning paths.
Through extensive experimental analysis, we have drawn several main findings: 
(1) HSP can guide LLMs to generate knowledge or key ideas to problems, thereby helping LLMs to generate more logically coherent reasoning paths to reach the correct answers (Sec.~\ref{sec:exp1_hsp}). 
(2) For the high-quality hint, the performance improvement of open-source models can reach 12.8, even surpassing ChatGPT (Sec.~\ref{sec:high-quality}).
(3) When meets challenging tasks, HSP fails on low-capability open-source LLMs (e.g., Llama2-7B); however, on high-capability open-source LLMs, under the self-consistency setting, HSP improves a lot on the samples with difficult topics or hard levels (Sec.~\ref{sec:exp4_hard_task}).
(4) Supervised fine-tuning on the GSM8K training dataset with the CoT+HSP format, our HSP-Llemma-7B (64.3) outperform GPT3.5 (57,1) and WizardMath-13B (63.9) (Sec.~\ref{sec:sft}).

\section*{Limitation}
Here, we summarize some limitations of this paper, as follows:
(1) The HSPMATH dataset was expanded by rewriting questions from GSM8K nine times, but our hints were generated based only on the original samples and applied to the nine rewritten samples. The rewritten samples might undergo logical changes, making the introduction of hints less harmonious. There might be a risk of poor performance during supervised fine-tuning. In the future, we will refine this dataset carefully and release a new version.
(2) Due to limitations in computational resources, this paper did not conduct supervised fine-tuning on models larger than 13B parameters in the SFT experiments, resulting in an incomplete exploration of HSP-enhanced supervised fine-tuning. We will undertake this exploration in the future.

\bibliography{custom}

\appendix
\newpage

\section{Robustness Analysis}
Considering the impact that varying sets of examples may have on results, the question arises: Is the HSP framework effective with diverse example sets?

To investigate this, we conducted experiments on the GSM8K (mathematical reasoning) and StrategyQA (common sense reasoning) datasets. Like the setting in Exp-I, we randomly chose four sets of examples from the testing set, each comprising $8$ examples for GSM8K and $6$ examples for StrategyQA. 
We then crafted hints and solutions featuring intermediate reasoning steps aided by GPT-4. 
These experiments were carried out on four LLMs: Llama2-7B, Llama2-13B, Llama2-70B, and Mixtral-8*7B. According to the results presented in Tab.~\ref{tab:robust}, CoT+HSP consistently outperformed CoT across the GSM8K and StrategyQA datasets, with all four models showing significant performance enhancements across the four example sets. This demonstrates the robustness of the performance gains achieved by integrating CoT with HSP.

\begin{table}[htbp]
  \centering \footnotesize
  \renewcommand\tabcolsep{2.1pt}
    \begin{tabular}{lccccccccc}
    \toprule
    \multirow{2}[2]{*}{Model} & \multirow{2}[2]{*}{HSP} & \multicolumn{4}{c}{GSM8K}     & \multicolumn{4}{c}{SQA} \\
    \cmidrule(lr){3-6} \cmidrule(lr){7-10}
          &       & E1    & E2    & E3    & E4    & E1    & E2    & E3    & E4 \\
    \midrule
    \multirow{2}[2]{*}{Lm2-7B} & $\times$ & 20.2  & 15.2  & 18.0  & 17.0  & 61.2  & 56.6  & 63.9  & 60.9 \\
          & $\checkmark$ & \textbf{22.7} & \textbf{21.6} & \textbf{23.4} & \textbf{22.8} & \textbf{63.8} & \textbf{61.5} & \textbf{65.9} & \textbf{63.3} \\
    \midrule
    \multirow{2}[2]{*}{Lm2-13B} & $\times$ & 35.9  & 29.1  & 25.4  & 32.2  & 64.1  & 60.6  & 67.5  & 63.2 \\
          & $\checkmark$ & \textbf{37.1} & \textbf{34.7} & \textbf{35.1} & \textbf{36.5} & \textbf{67.4} & \textbf{62.0} & \textbf{68.2} & \textbf{65.9} \\
    \midrule
    \multirow{2}[2]{*}{Lm2-70B} & $\times$ & 53.7  & 54.1  & 54.4  & 54.0  & 71.1  & 65.1  & 75.1  & 68.2 \\
          & $\checkmark$ & \textbf{60.1} & \textbf{56.3} & \textbf{55.3} & \textbf{59.3} & \textbf{71.7} & \textbf{72.1} & \textbf{75.8} & \textbf{73.1} \\
    \midrule
    \multirow{2}[2]{*}{Lm2-56B} & $\times$ & 67.9  & 68.8  & 67.2  & 67.8  & 65.4  & 60.3  & 69.3  & 61.9 \\
          & $\checkmark$ & \textbf{69.1} & \textbf{69.1} & \textbf{68.2} & \textbf{68.8} & \textbf{67.3} & \textbf{64.5} & \textbf{70.6} & \textbf{66.8} \\
    \bottomrule
    \end{tabular}%
      \caption{Experimental results for CoT Prompting with and without HSP on the GSM8K and StrategyQA (SQA) datasets across various example groups (E1, E2, E3, and E4). Values in bold denote the best results. 
      }
  \label{tab:robust}%
\end{table}%

\begin{table*}[htbp]
  \centering \footnotesize
    \begin{tabular}{p{38em}}
    \toprule
    \textbf{Mathematical reasoning } \\
    \midrule
    \textcolor{blue}{Please answer the following question.} \\
    \textcolor{blue}{Example 1:} 
    \textcolor{blue}{Question}: Shawn has five toys. For Christmas, he got two toys each from his mom and dad. How many toys does he have now?\\
    \textcolor{blue}{Hint}: Begin with the number of toys Shawn had initially. Then, add the number of toys he received from each parent. Remember, each parent gave him a certain number of toys, so you'll need to add those to his original amount to find out how many toys he has now. \\
   \textcolor{blue}{Solution}: Shawn started with 5 toys. If he got 2 toys each from his mom and dad, then that is 4 more toys. 5 + 4 = 9. The answer is 9.\\
   \textcolor{blue}{...... (Omitting 7 examples)} \\ \\
   \textcolor{blue}{Testing Example: } \\
    \textcolor{blue}{Question}: [QUESTION] \\
    \midrule
    \textbf{Commonsense reasoning} \\
    \midrule
    \textcolor{blue}{Please answer the following question.} \\
    \textcolor{blue}{Example 1:} 
    \textcolor{blue}{Question}: Do hamsters provide food for any animals? \\
    \textcolor{blue}{Hint}: Consider the natural role of hamsters in the food chain and who might rely on them as a source of nutrition.\\
    \textcolor{blue}{Solution}: Hamsters are prey animals. Prey are food for predators. Thus, hamsters provide food for some animals. So the answer is yes.\\
    \textcolor{blue}{...... (Omitting 5 examples)} \\ \\
   \textcolor{blue}{Testing Example: } \\ 
    \textcolor{blue}{Question}: [QUESTION] \\
    \bottomrule
    \end{tabular}%
    \vspace{-6pt}
    \caption{Prompt template for mathematical reasoning and commonsense reasoning.}
    \label{tab:prompt_template}
\end{table*}%

\section{Prompt Example}
The four models evaluated in this paper, namely Lm2-7B, Lm2-13B, Lm2-70B, and Mix-56B, were all tested using the same prompt template. Fig.~\ref{tab:prompt_template} shows the prompt template for mathematical reasoning and common sense reasoning tasks.

\section{Case Study}

Guiding the model to generate hints before the solution can effectively improve the model's performance. So, how does guiding LLM to generate hints first affect the generation of the model's solution?
We choose to introduce hints under CoT prompting and select case studies on mathematical reasoning and common sense reasoning tasks, as shown in Tab ~\ref{tab:case_study_all}.

\paragraph{Case 1}
The solution from CoT appears logical, but its analysis remains superficial, merely focusing on the relationship between the two entities (US brand Nice and the Western honey bee) to answer the question. In contrast, the hint from CoT+HSP suggests approaching from a deeper view, specifically questioning whether the crops relied upon by US brand Nice depend on Western honey bees for pollination, which leads to the correct answer.

\paragraph{Case 2}
For question ``Do black-tailed jackrabbits fear the European wildcat?'', CoT only considered the biological perspective, leading to an incorrect answer. However, the hint from CoT+HSP suggested that it is necessary to consider not only the biological aspect but also the habitat of the organism, thereby achieving the correct answer.

\paragraph{Case 3}
We can observe that CoT's calculation method overlooks an important piece of knowledge, namely the formula for calculating the perimeter: ``The distance traveled by a point on the edge of a rotating object equals the circle's circumference.''. In contrast, CoT+Hint successfully suggests utilizing the formula for perimeter, thereby obtaining the correct answer.

\paragraph{Case 4}
The question involves calculating the perimeter of a rectangle, but the CoT method only adds the width and height of a rectangle. CoT+HSP suggested that the perimeter be calculated by four lengths, making the final answer calculation correct.

\begin{table*}[htb]
  \centering \footnotesize
    \begin{tabular}{p{49em}}
    \toprule
    \textbf{Case 1: Commonsense reasoning} \\
    \midrule
    \textcolor{blue}{Question}: Does US brand Nice depend on Western honey bee? \\
    \textcolor{blue}{True Answer}: TRUE \\
    \textcolor{blue}{CoT Response}: Nice is a brand of ice cream. Ice cream is not made from honey bees. Thus, Nice does not depend on Western honey bees. So the answer is no. \textcolor{cyan}{[WRONG]}\\
    \textcolor{blue}{CoT+HSP Response}: \\\textcolor{redpink}{Hint: Consider the crops that Nice brand depends on and whether those crops require pollination by Western honey bees.} \\Solution: Nice brand depends on almonds. Almonds require pollination by Western honey bees. Thus, Nice brand depends on Western honey bees. 
    So the answer is yes. \textcolor{cyan}{[CORRECT]}\\
    \midrule
    \textbf{Case 2: Commonsense reasoning} \\
    \midrule
    \textcolor{blue}{Question}: Do black-tailed jackrabbits fear the European wildcat? \\
    \textcolor{blue}{True Answer}: FALSE \\
    \textcolor{blue}{CoT Response}: Black-tailed jackrabbits are prey animals. European wildcats are predators. Thus, black-tailed jackrabbits would fear the European wildcat. So the answer is yes. \textcolor{cyan}{[WRONG]}\\
    \textcolor{blue}{CoT+HSP Response}: \\\textcolor{redpink}{Hint: Consider the natural habitats of black-tailed jackrabbits and European wildcats and whether their paths would cross.} \\Solution: Black-tailed jackrabbits are native to North America. European wildcats are native to Europe. Thus, their paths would not naturally cross. So the answer is no. \textcolor{cyan}{[CORRECT]}\\
    \midrule
    \textbf{Case 3: Mathematical reasoning} \\
    \midrule
    \textcolor{blue}{Question}: A point on the edge of a fan blade that is rotating in a plane 10 centimeters from the center of the fan. What is the distance traveled, in centimeters, by this point after 30 seconds when the fan runs at the rate of 300 revolutions per minutes? \# Answer option: [A)$750\pi$, B)$1500\pi$, C) $1875\pi$, D)$3000\pi$, E)$7500\pi$] \\
    \textcolor{blue}{True Answer}: D \\
    \textcolor{blue}{CoT Response}: The fan makes 300 revolutions in 60 seconds. So it makes 300/60 = 5 revolutions in 1 second. So it makes 5*30 = 150 revolutions in 30 seconds. The point on the edge of the fan blade travels 10 centimeters in one revolution. So it travels 10*150 = 1500 centimeters in 30 seconds. So the answer is (b). \textcolor{cyan}{[WRONG]}\\
    \textcolor{blue}{CoT+HSP Response}: \\\textcolor{redpink}{Hint: The distance traveled by a point on the edge of a rotating object is calculated by multiplying the circumference of the circle it moves along by the number of rotations.} \\Solution: The fan makes 300 revolutions in 60 seconds. In 30 seconds, it makes 300/2 = 150 revolutions. The circumference of the circle is $2\pi r = 2\pi(10) = 20\pi$. The distance traveled is $150 * 20\pi = 3000 \pi$. So the answer is (d). \textcolor{cyan}{[CORRECT]}\\
    \midrule
    \textbf{Case 4: Mathematical  reasoning} \\
    \midrule
    \textcolor{blue}{Question}: James wants to hang crepe-paper streamers from his backyard fence for his birthday party. His backyard fence is a rectangle that measures 20 feet on the long side and 15 feet on the short side. How many feet of crepe paper does James need to buy? \\
    \textcolor{blue}{True Answer}: 70 \\
    \textcolor{blue}{CoT Response}: The fence is 20 feet long and 15 feet wide. The total length of the fence is 20 + 15 = 35 feet. He needs exactly the same amount of crepe paper as the length of the fence. So he needs to buy 35 feet of crepe paper. The answer is 35. \textcolor{cyan}{[WRONG]}\\
    \textcolor{blue}{CoT+HSP Response}: \\\textcolor{redpink}{Hint: To find out how much crepe paper James needs to buy, calculate the perimeter of the fence. The perimeter is the total distance around the fence. To find the perimeter, add the lengths of all four sides of the fence together.} \\Solution: The long side of the fence is 20 feet. The short side is 15 feet. So the perimeter is 20 + 15 + 20 + 15 = 70 feet. The answer is 70. \textcolor{cyan}{[CORRECT]}\\
    \bottomrule
    \end{tabular}%
    \vspace{-6pt}
    \caption{Case studies of solving mathematical reasoning and commonsense reasoning problems with CoT+HSP and CoT prompting on the Mixtral-7*8B model. \textcolor{blue}{Blue} text indicates the stem, \textcolor{redpink}{pink} text indicates the effective hint, \textcolor{cyan}{cyan} text indicates the judgment of whether the answer is correct, \textcolor{cyan}{[CORRECT]} denotes correct, and \textcolor{cyan}{[WRONG]} denotes incorrect.}
   \label{tab:case_study_all}%
\end{table*}%

\section{Reference Baseline}
\label{sec:baseline_result}
\begin{figure*}[ht]
	\centering
        \subfigure[GSM8K]{
	\begin{minipage}[b]{0.48 \textwidth}
		\includegraphics[width=1\textwidth]{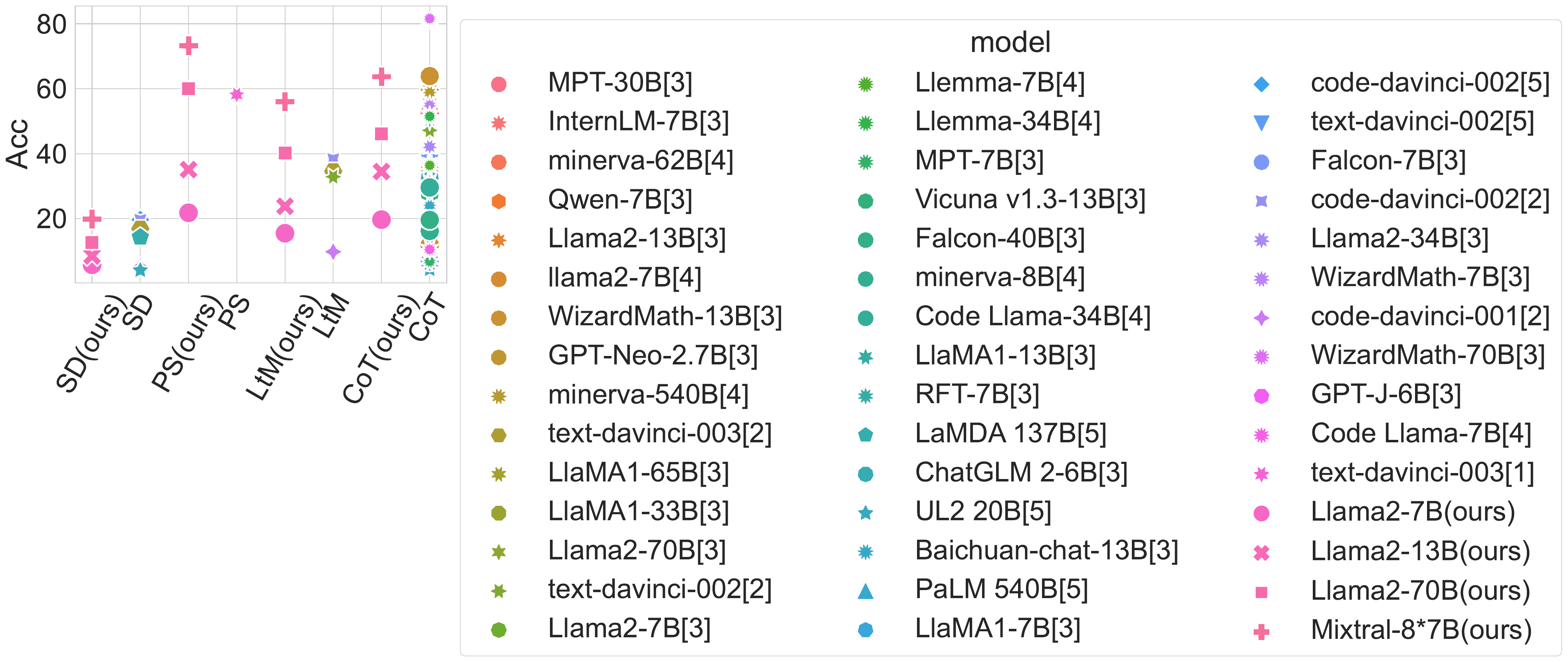}
	\end{minipage}
	} 
     \subfigure[ASDiv]{
	\begin{minipage}[b]{0.48 \textwidth}
		\includegraphics[width=1\textwidth]{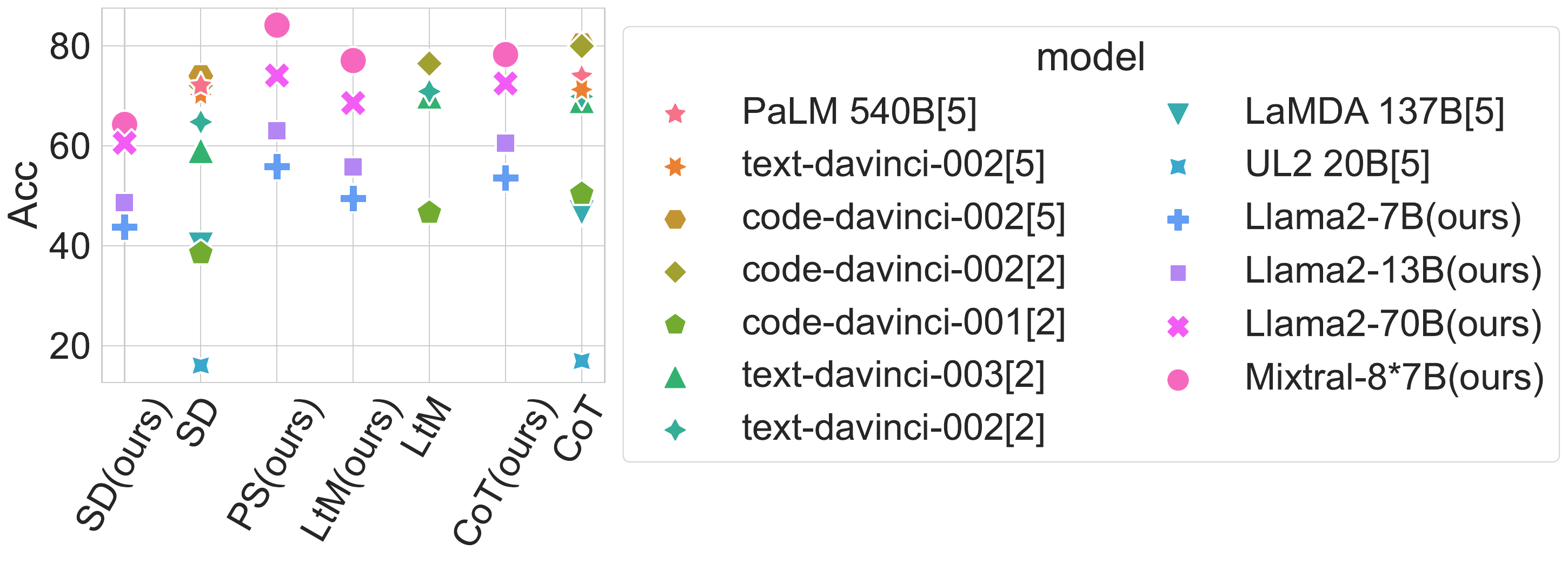}
	\end{minipage}
	}  \\
	
        \subfigure[MultiArith]{
	\begin{minipage}[b]{0.48 \textwidth}
		\includegraphics[width=1\textwidth]{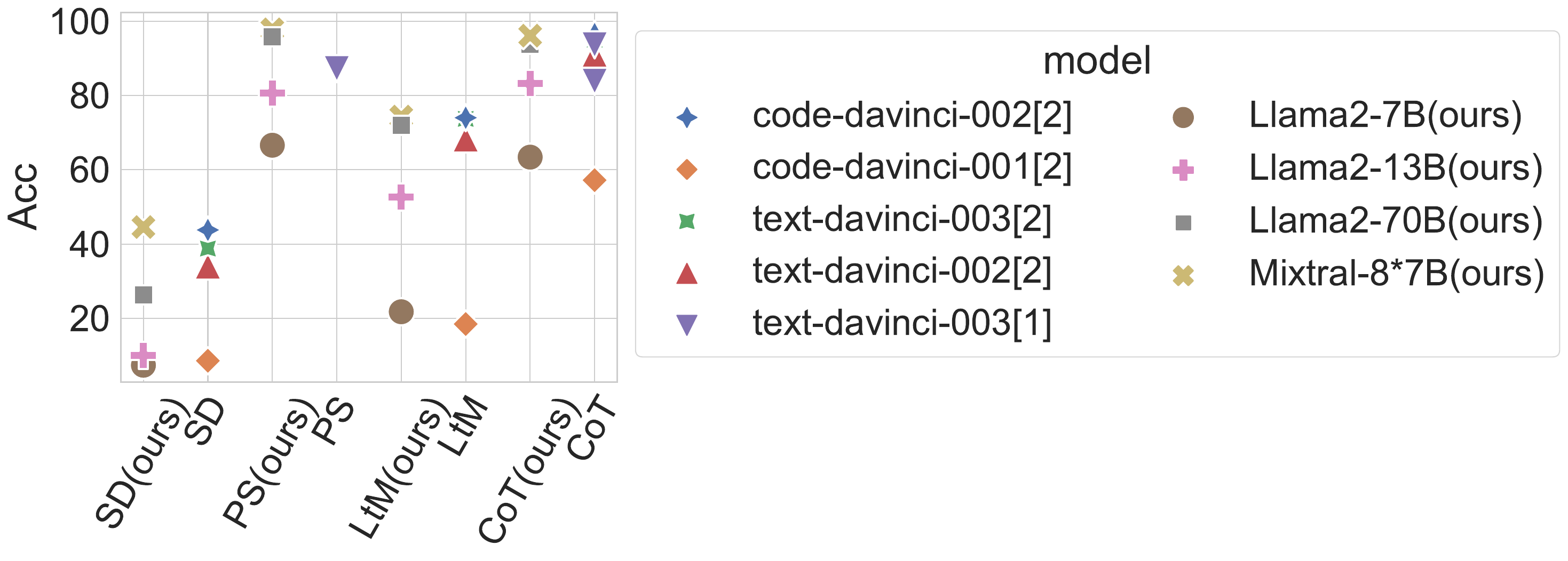}
	\end{minipage}
	} 
        \subfigure[AQUA]{
	\begin{minipage}[b]{0.48 \textwidth}
		\includegraphics[width=1\textwidth]{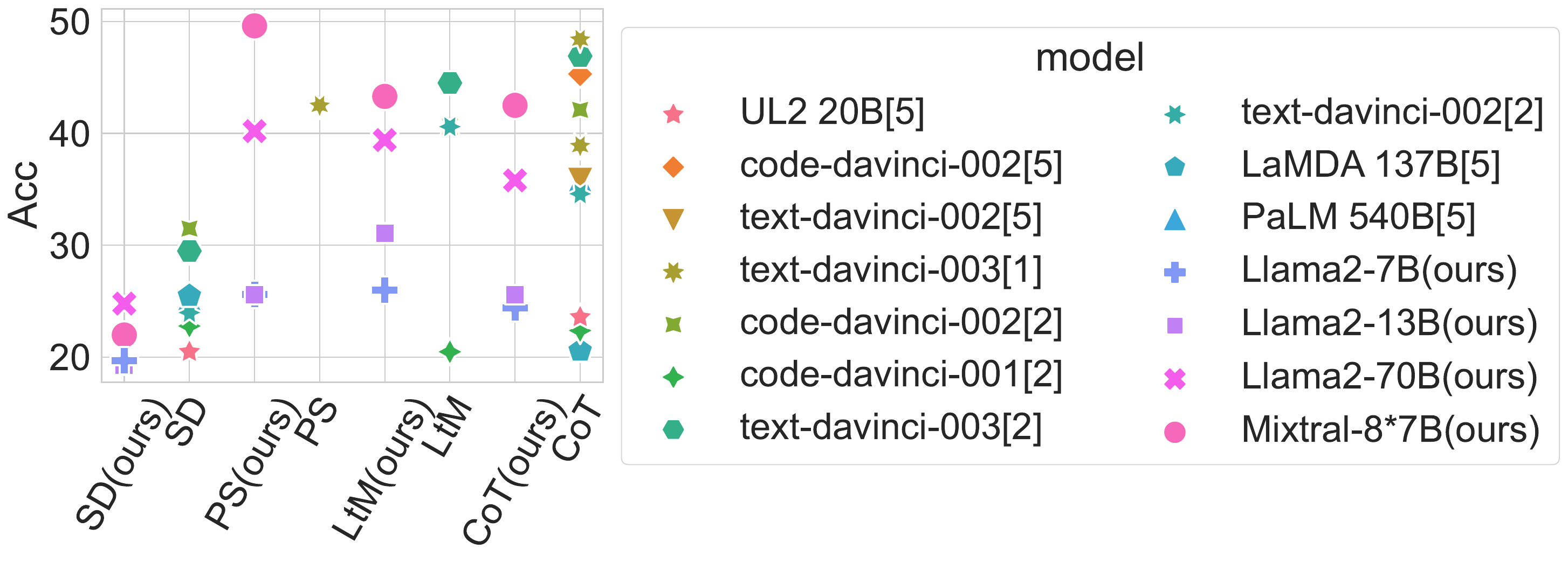}
	\end{minipage}
	}  \\
        \subfigure[StrategyQA]{
	\begin{minipage}[b]{0.48 \textwidth}
		\includegraphics[width=1\textwidth]{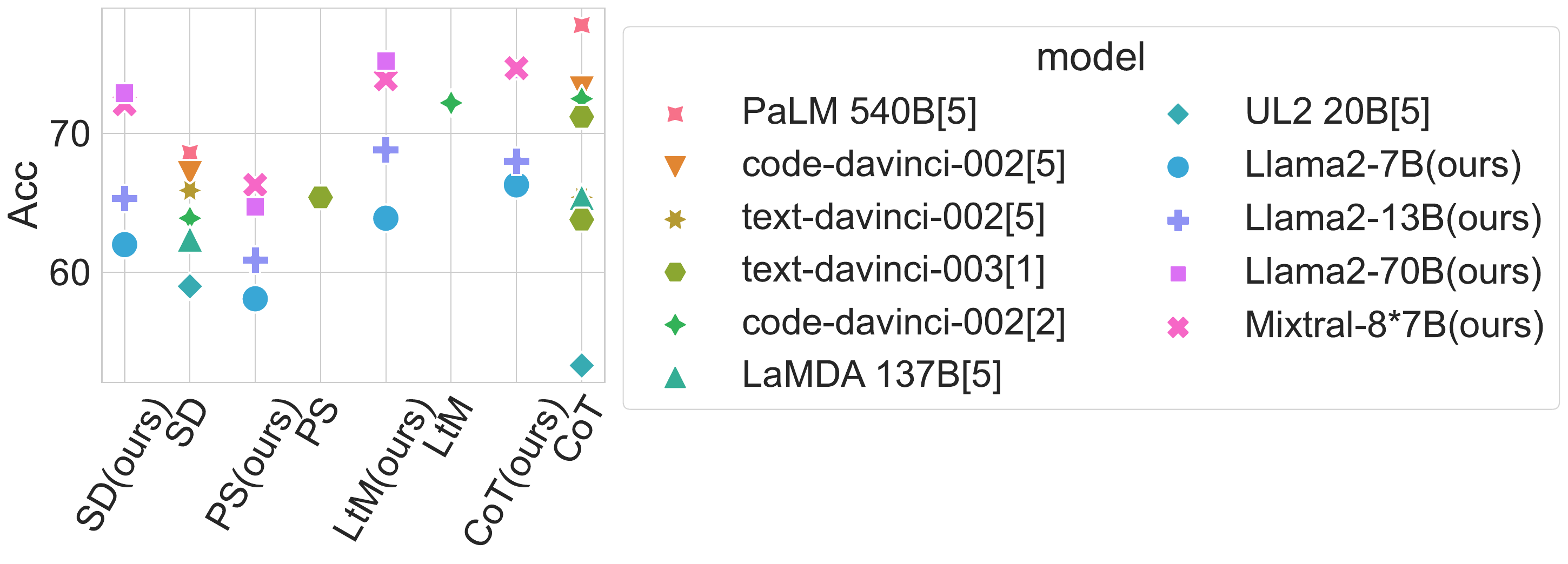}
	\end{minipage}
	} 
	\subfigure[Date]{
	\begin{minipage}[b]{0.48 \textwidth}
		\includegraphics[width=1\textwidth]{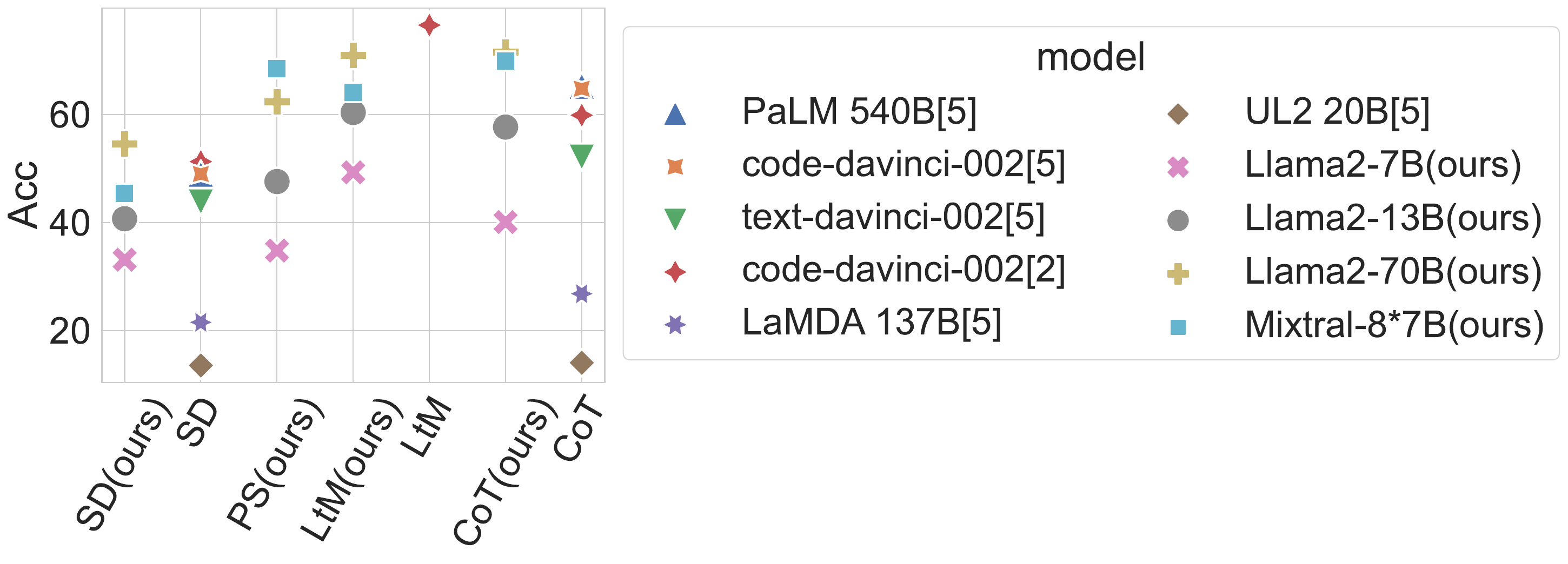}
	\end{minipage}
	} 
	\vspace{-6pt}
	\caption{A comparison of the results from existing work with the results reimplemented in this work for Llama2-7B, Llama2-13B, Llama2-70B, and Mixtral-7*8B across six datasets. The existing results come from five works: [1]~\cite{wang:psp}, [2]~\cite{faithfulcot:lyu}, [3]~\cite{wizardmath:luo}, [4]~\cite{Llemma}, and [5]~\cite{wei2022chain}.}
\label{fig:reference_baseline}
\end{figure*}

In this paper, we reimplemented the results of four models, namely Llama-7B, Llama-13B, Llama-70B, and Mixtral-7*8B, under SD, LtM, PS, and CoT promptings, to compare with our HSP-enhanced promptings' performance. Are our reimplemented results within a reasonable range?
To answer this question, we compared our reimplemented results with results from some recently works across six datasets: GSM8K, AQUA, ASDiv, Date, MultiArith, and StrategyQA.
The results are shown in Fig.~\ref{fig:reference_baseline}.

There is a considerable amount of existing work on CoT prompting, while results for SD, LtM, and PS prompting are limit.
The baseline work we present in the Fig.~\ref{fig:reference_baseline} comes from five studies that cover a broad range of baseline methods.
We can observe that across these six datasets, except for Llama-7B, which often lacks a closely matched model size for a baseline, the results for Llama-13B, Llama-70B, and Mixtral-7*8B are comparable to some existing open-source or closed-source models.

\section{Results of Self-consistency}
\label{sec:sc_full_appendix}
Tab.~\ref{tab:sc_full} shows the results of self-consistency.

\begin{table*}[htb]
  \centering \footnotesize
  \renewcommand\tabcolsep{3.5pt}
    \begin{tabular}{ccccccccccccc}
    \toprule
    \multirow{2}[2]{*}{Model} & \multirow{2}[2]{*}{SC} & \multirow{2}[2]{*}{Hint} & \multicolumn{4}{c}{MATH}      & \multicolumn{2}{c}{Commonsense} & \multirow{2}[2]{*}{Avg} & \multicolumn{3}{c}{\multirow{2}[2]{*}{Relative Improvement}} \\
          &       &       & GSM8K & ASDiv & MultiArith & AQUA  & SQA   & Date  &       & \multicolumn{3}{c}{} \\
    \midrule
    \multirow{15}[2]{*}{Llama2-7B} & 1     & 0     & 19.7  & 53.6  & 63.4  & 24.4  & 66.3  & \textcolor[rgb]{ .898,  .263,  .875}{\textbf{40.1}} & 44.6  & \multicolumn{3}{c}{\multirow{15}[2]{*}{\includegraphics[scale=0.5]{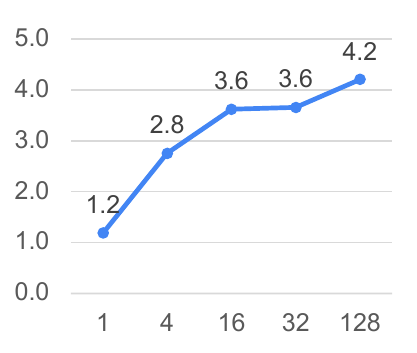}}} \\
          & 1     & 1     & 19.9  & 55.8  & 63.8  & 24.4  & 67.5  & 43.2  & 45.8  & \multicolumn{3}{c}{} \\
          & \textcolor[rgb]{ 0,  .69,  .314}{1} & \textcolor[rgb]{ 0,  .69,  .314}{Impv} & \textcolor[rgb]{ 0,  .69,  .314}{0.2} & \textcolor[rgb]{ 0,  .69,  .314}{2.2} & \textcolor[rgb]{ 0,  .69,  .314}{0.4} & \textcolor[rgb]{ 0,  .69,  .314}{0.0} & \textcolor[rgb]{ 0,  .69,  .314}{1.2} & \textcolor[rgb]{ 0,  .69,  .314}{3.1} & \textcolor[rgb]{ 0,  .69,  .314}{1.2} & \multicolumn{3}{c}{} \\
          & 4     & 0     & 23.6  & 54.6  & 68.0  & 23.6  & 67.9  & \textcolor[rgb]{ .898,  .263,  .875}{\textbf{40.1}} & 46.3  & \multicolumn{3}{c}{} \\
          & 4     & 1     & 26.5  & 57.1  & 73.0  & 26.4  & 69.2  & 42.1  & 49.1  & \multicolumn{3}{c}{} \\
          & 4     & \textcolor[rgb]{ 0,  .69,  .314}{Impv} & \textcolor[rgb]{ 0,  .69,  .314}{2.9} & \textcolor[rgb]{ 0,  .69,  .314}{2.5} & \textcolor[rgb]{ 0,  .69,  .314}{5.0} & \textcolor[rgb]{ 0,  .69,  .314}{2.8} & \textcolor[rgb]{ 0,  .69,  .314}{1.3} & \textcolor[rgb]{ 0,  .69,  .314}{2.0} & \textcolor[rgb]{ 0,  .69,  .314}{2.8} & \multicolumn{3}{c}{} \\
          & 16    & 0     & 24.7  & \textcolor[rgb]{ .898,  .263,  .875}{\textbf{55.5}} & \textcolor[rgb]{ .898,  .263,  .875}{\textbf{68.5}} & 26.0  & 67.9  & \textcolor[rgb]{ .898,  .263,  .875}{\textbf{40.1}} & 47.1  & \multicolumn{3}{c}{} \\
          & 16    & 1     & 29.2  & 57.3  & 77.7  & 26.0  & \textcolor[rgb]{ .259,  .522,  .957}{\textbf{70.7}} & 43.5  & 50.7  & \multicolumn{3}{c}{} \\
          & 16    & \textcolor[rgb]{ 0,  .69,  .314}{Impv} & \textcolor[rgb]{ 0,  .69,  .314}{4.5} & \textcolor[rgb]{ 0,  .69,  .314}{1.8} & \textcolor[rgb]{ 0,  .69,  .314}{9.2} & \textcolor[rgb]{ 0,  .69,  .314}{0.0} & \textcolor[rgb]{ 0,  .69,  .314}{2.8} & \textcolor[rgb]{ 0,  .69,  .314}{3.4} & \textcolor[rgb]{ 0,  .69,  .314}{3.6} & \multicolumn{3}{c}{} \\
          & 32    & 0     & \textcolor[rgb]{ .898,  .263,  .875}{25.5} & 55.2  & 67.6  & 25.6  & \textcolor[rgb]{ .898,  .263,  .875}{\textbf{68.6}} & 39.6  & 47.0  & \multicolumn{3}{c}{} \\
          & 32    & 1     & 29.5  & 57.5  & 78.9  & \textcolor[rgb]{ .259,  .522,  .957}{\textbf{26.4}} & 70.2  & 41.5  & 50.7  & \multicolumn{3}{c}{} \\
          & 32    & \textcolor[rgb]{ 0,  .69,  .314}{Impv} & \textcolor[rgb]{ 0,  .69,  .314}{4.0} & \textcolor[rgb]{ 0,  .69,  .314}{2.3} & \textcolor[rgb]{ 0,  .69,  .314}{11.3} & \textcolor[rgb]{ 0,  .69,  .314}{0.8} & \textcolor[rgb]{ 0,  .69,  .314}{1.6} & \textcolor[rgb]{ 0,  .69,  .314}{1.9} & \textcolor[rgb]{ 0,  .69,  .314}{3.6} & \multicolumn{3}{c}{} \\
          & 128   & 0     & 25.4  & 55.1  & 68.1  & \textcolor[rgb]{ .898,  .263,  .875}{\textbf{26.4}} & 68.3  & 40.4  & 47.3  & \multicolumn{3}{c}{} \\
          & 128   & 1     & \textcolor[rgb]{ .259,  .522,  .957}{\textbf{30.3}} & \textcolor[rgb]{ .259,  .522,  .957}{\textbf{59.0}} & \textcolor[rgb]{ .259,  .522,  .957}{\textbf{79.5}} & 25.6  & 70.2  & \textcolor[rgb]{ .259,  .522,  .957}{\textbf{44.3}} & 51.5  & \multicolumn{3}{c}{} \\
          & 128   & \textcolor[rgb]{ 0,  .69,  .314}{Impv} & \textcolor[rgb]{ 0,  .69,  .314}{4.9} & \textcolor[rgb]{ 0,  .69,  .314}{3.9} & \textcolor[rgb]{ 0,  .69,  .314}{11.4} & \textcolor[rgb]{ 0,  .69,  .314}{-0.8} & \textcolor[rgb]{ 0,  .69,  .314}{1.9} & \textcolor[rgb]{ 0,  .69,  .314}{3.9} & \textcolor[rgb]{ 0,  .69,  .314}{4.2} & \multicolumn{3}{c}{} \\
    \midrule
    \multirow{15}[2]{*}{Llama2-13B} & 1     & 0     & 34.5  & 60.5  & 83.2  & 25.6  & 68.0  & 52.4  & 54.0  & \multicolumn{3}{c}{\multirow{15}[2]{*}{\includegraphics[scale=0.5]{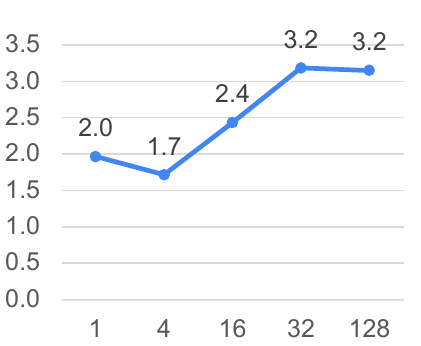}}} \\
          & 1     & 1     & 36.5  & 61.2  & 87.1  & 25.6  & 72.1  & 53.5  & 56.0  & \multicolumn{3}{c}{} \\
          & 1     & \textcolor[rgb]{ 0,  .69,  .314}{Impv} & \textcolor[rgb]{ 0,  .69,  .314}{2.0} & \textcolor[rgb]{ 0,  .69,  .314}{0.7} & \textcolor[rgb]{ 0,  .69,  .314}{3.9} & \textcolor[rgb]{ 0,  .69,  .314}{0.0} & \textcolor[rgb]{ 0,  .69,  .314}{4.1} & \textcolor[rgb]{ 0,  .69,  .314}{1.1} & \textcolor[rgb]{ 0,  .69,  .314}{2.0} & \multicolumn{3}{c}{} \\
          & 4     & 0     & 40.7  & 61.5  & 87.8  & 25.6  & 69.1  & 57.4  & 57.0  & \multicolumn{3}{c}{} \\
          & 4     & 1     & 41.1  & 62.7  & 89.4  & 28.7  & 72.6  & 57.9  & 58.7  & \multicolumn{3}{c}{} \\
          & 4     & \textcolor[rgb]{ 0,  .69,  .314}{Impv} & \textcolor[rgb]{ 0,  .69,  .314}{0.4} & \textcolor[rgb]{ 0,  .69,  .314}{1.2} & \textcolor[rgb]{ 0,  .69,  .314}{1.6} & \textcolor[rgb]{ 0,  .69,  .314}{3.1} & \textcolor[rgb]{ 0,  .69,  .314}{3.5} & \textcolor[rgb]{ 0,  .69,  .314}{0.5} & \textcolor[rgb]{ 0,  .69,  .314}{1.7} & \multicolumn{3}{c}{} \\
          & 16    & 0     & 42.3  & 62.5  & 89.4  & \textcolor[rgb]{ .898,  .263,  .875}{\textbf{28.0}} & 69.0  & 57.7  & 58.2  & \multicolumn{3}{c}{} \\
          & 16    & 1     & 46.9  & 64.7  & 91.3  & 28.7  & 72.8  & 59.1  & 60.6  & \multicolumn{3}{c}{} \\
          & 16    & \textcolor[rgb]{ 0,  .69,  .314}{Impv} & \textcolor[rgb]{ 0,  .69,  .314}{4.6} & \textcolor[rgb]{ 0,  .69,  .314}{2.2} & \textcolor[rgb]{ 0,  .69,  .314}{1.9} & \textcolor[rgb]{ 0,  .69,  .314}{0.7} & \textcolor[rgb]{ 0,  .69,  .314}{3.8} & \textcolor[rgb]{ 0,  .69,  .314}{1.4} & \textcolor[rgb]{ 0,  .69,  .314}{2.4} & \multicolumn{3}{c}{} \\
          & 32    & 0     & 41.5  & 62.6  & 90.1  & 26.4  & 69.6  & 57.9  & 58.0  & \multicolumn{3}{c}{} \\
          & 32    & 1     & 48.2  & 64.9  & 92.8  & 28.0  & \textcolor[rgb]{ .259,  .522,  .957}{\textbf{73.1}} & 60.2  & 61.2  & \multicolumn{3}{c}{} \\
          & 32    & \textcolor[rgb]{ 0,  .69,  .314}{Impv} & \textcolor[rgb]{ 0,  .69,  .314}{6.7} & \textcolor[rgb]{ 0,  .69,  .314}{2.3} & \textcolor[rgb]{ 0,  .69,  .314}{2.7} & \textcolor[rgb]{ 0,  .69,  .314}{1.6} & \textcolor[rgb]{ 0,  .69,  .314}{3.5} & \textcolor[rgb]{ 0,  .69,  .314}{2.3} & \textcolor[rgb]{ 0,  .69,  .314}{3.2} & \multicolumn{3}{c}{} \\
          & 128   & 0     & \textcolor[rgb]{ .898,  .263,  .875}{\textbf{47.9}} & \textcolor[rgb]{ .898,  .263,  .875}{\textbf{62.9}} & \textcolor[rgb]{ .898,  .263,  .875}{\textbf{90.1}} & 27.6  & \textcolor[rgb]{ .898,  .263,  .875}{\textbf{70.2}} & \textcolor[rgb]{ .898,  .263,  .875}{\textbf{58.8}} & 59.6  & \multicolumn{3}{c}{} \\
          & 128   & 1     & \textcolor[rgb]{ .259,  .522,  .957}{\textbf{52.5}} & \textcolor[rgb]{ .259,  .522,  .957}{\textbf{66.8}} & \textcolor[rgb]{ .259,  .522,  .957}{\textbf{93.0}} & \textcolor[rgb]{ .259,  .522,  .957}{\textbf{29.5}} & 73.0  & \textcolor[rgb]{ .259,  .522,  .957}{\textbf{61.6}} & 62.7  & \multicolumn{3}{c}{} \\
          & 128   & \textcolor[rgb]{ 0,  .69,  .314}{Impv} & \textcolor[rgb]{ 0,  .69,  .314}{4.6} & \textcolor[rgb]{ 0,  .69,  .314}{3.9} & \textcolor[rgb]{ 0,  .69,  .314}{2.9} & \textcolor[rgb]{ 0,  .69,  .314}{1.9} & \textcolor[rgb]{ 0,  .69,  .314}{2.8} & \textcolor[rgb]{ 0,  .69,  .314}{2.8} & \textcolor[rgb]{ 0,  .69,  .314}{3.2} & \multicolumn{3}{c}{} \\
    \midrule
    \multirow{15}[2]{*}{Llama2-70B} & 1     & 0     & 46.1  & 72.5  & 93.8  & 35.8  & 74.6  & 71.6  & 65.7  & \multicolumn{3}{c}{\multirow{15}[2]{*}{\includegraphics[scale=0.5]{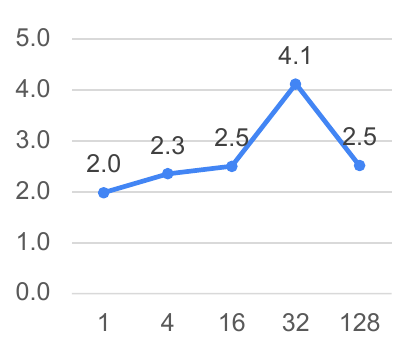}}} \\
          & 1     & 1     & 50.3  & 74.4  & 94.6  & 37.0  & 77.0  & 73.0  & 67.7  & \multicolumn{3}{c}{} \\
          & 1     & \textcolor[rgb]{ 0,  .69,  .314}{Impv} & \textcolor[rgb]{ 0,  .69,  .314}{4.2} & \textcolor[rgb]{ 0,  .69,  .314}{1.9} & \textcolor[rgb]{ 0,  .69,  .314}{0.8} & \textcolor[rgb]{ 0,  .69,  .314}{1.2} & \textcolor[rgb]{ 0,  .69,  .314}{2.4} & \textcolor[rgb]{ 0,  .69,  .314}{1.4} & \textcolor[rgb]{ 0,  .69,  .314}{2.0} & \multicolumn{3}{c}{} \\
          & 4     & 0     & 59.5  & 75.0  & 95.3  & 39.8  & 78.2  & 73.3  & 70.2  & \multicolumn{3}{c}{} \\
          & 4     & 1     & 60.5  & 75.9  & 96.1  & 41.3  & 78.3  & 82.7  & 72.5  & \multicolumn{3}{c}{} \\
          & 4     & \textcolor[rgb]{ 0,  .69,  .314}{Impv} & \textcolor[rgb]{ 0,  .69,  .314}{1.0} & \textcolor[rgb]{ 0,  .69,  .314}{0.9} & \textcolor[rgb]{ 0,  .69,  .314}{0.8} & \textcolor[rgb]{ 0,  .69,  .314}{1.5} & \textcolor[rgb]{ 0,  .69,  .314}{0.1} & \textcolor[rgb]{ 0,  .69,  .314}{9.4} & \textcolor[rgb]{ 0,  .69,  .314}{2.3} & \multicolumn{3}{c}{} \\
          & 16    & 0     & 60.1  & 76.3  & 96.1  & 42.1  & 78.4  & 73.3  & 71.1  & \multicolumn{3}{c}{} \\
          & 16    & 1     & 67.0  & 77.9  & 97.8  & 44.5  & 79.2  & 74.9  & 73.6  & \multicolumn{3}{c}{} \\
          & 16    & \textcolor[rgb]{ 0,  .69,  .314}{Impv} & \textcolor[rgb]{ 0,  .69,  .314}{6.9} & \textcolor[rgb]{ 0,  .69,  .314}{1.6} & \textcolor[rgb]{ 0,  .69,  .314}{1.7} & \textcolor[rgb]{ 0,  .69,  .314}{2.4} & \textcolor[rgb]{ 0,  .69,  .314}{0.8} & \textcolor[rgb]{ 0,  .69,  .314}{1.6} & \textcolor[rgb]{ 0,  .69,  .314}{2.5} & \multicolumn{3}{c}{} \\
          & 32    & 0     & 60.6  & 77.1  & 96.3  & 45.3  & 78.5  & 72.7  & 71.8  & \multicolumn{3}{c}{} \\
          & 32    & 1     & 67.4  & 78.4  & 98.0  & \textcolor[rgb]{ .361,  .541,  .984}{\textbf{47.2}} & \textcolor[rgb]{ .361,  .541,  .984}{\textbf{80.1}} & \textcolor[rgb]{ .361,  .541,  .984}{\textbf{84.1}} & 75.9  & \multicolumn{3}{c}{} \\
          & 32    & \textcolor[rgb]{ 0,  .69,  .314}{Impv} & \textcolor[rgb]{ 0,  .69,  .314}{6.8} & \textcolor[rgb]{ 0,  .69,  .314}{1.3} & \textcolor[rgb]{ 0,  .69,  .314}{1.7} & \textcolor[rgb]{ 0,  .69,  .314}{1.9} & \textcolor[rgb]{ 0,  .69,  .314}{1.6} & \textcolor[rgb]{ 0,  .69,  .314}{11.4} & \textcolor[rgb]{ 0,  .69,  .314}{4.1} & \multicolumn{3}{c}{} \\
          & 128   & 0     & \textcolor[rgb]{ .898,  .263,  .875}{\textbf{67.0}} & \textcolor[rgb]{ .898,  .263,  .875}{\textbf{77.6}} & \textcolor[rgb]{ .898,  .263,  .875}{\textbf{96.3}} & \textcolor[rgb]{ .898,  .263,  .875}{\textbf{46.1}} & \textcolor[rgb]{ .898,  .263,  .875}{\textbf{79.4}} & \textcolor[rgb]{ .898,  .263,  .875}{\textbf{73.5}} & 73.3  & \multicolumn{3}{c}{} \\
          & 128   & 1     & \textcolor[rgb]{ .361,  .541,  .984}{\textbf{67.6}} & \textcolor[rgb]{ .361,  .541,  .984}{\textbf{78.8}} & \textcolor[rgb]{ .361,  .541,  .984}{\textbf{98.2}} & 47.6  & 79.5  & 83.3  & 75.8  & \multicolumn{3}{c}{} \\
          & 128   & \textcolor[rgb]{ 0,  .69,  .314}{Impv} & \textcolor[rgb]{ 0,  .69,  .314}{0.6} & \textcolor[rgb]{ 0,  .69,  .314}{1.2} & \textcolor[rgb]{ 0,  .69,  .314}{1.9} & \textcolor[rgb]{ 0,  .69,  .314}{1.5} & \textcolor[rgb]{ 0,  .69,  .314}{0.1} & \textcolor[rgb]{ 0,  .69,  .314}{9.8} & \textcolor[rgb]{ 0,  .69,  .314}{2.5} & \multicolumn{3}{c}{} \\
    \bottomrule
    \end{tabular}%
      \caption{The results of self-consistency on the six datasets. Values in green denote the relative performance improvement with hints versus without hints under the same setting. The blue bold values represent the best performance with hints, while the pink bold values indicate the best performance without hints. The figure on the right shows the average relative improvement across six datasets.}
  \label{tab:sc_full}%
\end{table*}%

\end{document}